\documentclass[10pt,twocolumn,letterpaper]{article}

\usepackage[pagenumbers]{cvpr} % To force page numbers, e.g. for an arXiv version

\definecolor{cvprblue}{rgb}{0.21,0.49,0.74}
\usepackage[pagebackref,breaklinks,colorlinks,allcolors=cvprblue]{hyperref}
\usepackage{graphicx}    % 用于插入图片
\usepackage{caption}     % 用于配置标题
\usepackage{subcaption}  % 用于创建子图
\usepackage{geometry}    % 可选，用于调整页面边距
\usepackage{multirow}
\usepackage{tabularx}
\usepackage{colortbl}
\usepackage{xcolor}
\usepackage{multirow}
\usepackage{booktabs}

\usepackage{graphicx}

\usepackage{booktabs}    % For \toprule, \midrule, \bottomrule
\usepackage{xcolor}      % For \textcolor
\usepackage{array}        % For extended column definitions if needed

\usepackage{ifpdf}
\title{Are Conditional Latent Diffusion Models Effective for Image Restoration?}
\author{Yunchen Yuan$^{1}$,  Junyuan Xiao$^{2}$,  Xinjie Li$^{3,\dag}$ \\ 
\small{$^1$The Hong Kong Polytechnic University}\quad
\small{$^2$Tsinghua University}\quad
\small{$^3$Pennsylvania State University}\quad
}

\begin{document}

\setlength{\abovedisplayskip}{3pt}
\setlength{\belowdisplayskip}{2pt}

\maketitle
% \input{sec/0_abstract}

% Recently, there has been a surge in the development of Image Restoration (IR) based on Conditional Latent Diffusion Models (CLDM). Despite the increasing performance observed over the past few years, this work questions the validity of this research trajectory.
% Conditional latent diffusion models are arguably a successful approach for extracting high-level semantic correlations between images and conditions, facilitating text-to-image generation with spatial conditioning control. However, in the context of image restoration, the objective is to enhance the perceptual quality of images while maintaining semantic consistency. The semantic-level control characteristics of the CLDMs led to the difficulty of modeling the relationship between degraded images and ground truth images using a low-level representation. To substantiate our claims, we conducted experiments utilizing several traditional IR models alongside state-of-the-art CLDMs for comparative analysis. The experimental results indicate that, despite their significant scaling advantages, CLDMs exhibit issues such as high distortion and semantic deviation, particularly in samples with low levels of degradation, where traditional methods demonstrate superior performance. Furthermore, we conduct empirical studies to investigate the effects of various design elements of CLDMs on the performance. We anticipate this unexpected finding will pave the way for new research directions within the IR task. 

\begin{abstract}

Recent advancements in image restoration increasingly employ conditional latent diffusion models (CLDMs). While these models have demonstrated notable performance improvements in recent years, this work questions their suitability for IR tasks.
CLDMs excel in capturing high-level semantic correlations, making them effective for tasks like text-to-image generation with spatial conditioning. However, in IR, where the goal is to enhance image perceptual quality, these models face difficulty of modeling the relationship between degraded images and ground truth images using a low-level representation. 
To support our claims, we compare state-of-the-art CLDMs with traditional image restoration models through extensive experiments. Results reveal that despite the scaling advantages of CLDMs, they suffer from high distortion and semantic deviation, especially in cases with minimal degradation, where traditional methods outperform them. Additionally, we perform empirical studies to examine the impact of various CLDM design elements on their restoration performance. We hope this finding inspires a reexamination of current CLDM-based IR solutions, opening up more opportunities in this field.

\end{abstract}

\section{Introduction}

Image Restoration (IR) is a cornerstone of low-level vision research\cite{wang2020deep, wang2024exploiting, yang2023pasd, zhang2021designing, chen2022real,li2022survey, tao2018scale, whang2022deblurring, ren2023multiscale, tsai2022stripformer, li2023realworlddeeplocalmotion,Jebur2024,li2021low}, playing a crucial role in improving the visual quality of images.
Traditional IR solutions rely on task-specific knowledge to model degradation and restore images using mathematical modeling and classical signal processing algorithms~\cite{amudha2012survey}. Over the past few decades, IR approaches have evolved toward deep learning-based methods, leveraging neural networks to model and enhance the restoration process~\cite{su2022survey}.

In recent years, diffusion models (DMs) have gained prominence as state-of-the-art image generation models, celebrated for their powerful generative capabilities and adaptability~\cite{ho2020denoising, rombach2022high}. Conditional Latent Diffusion Models (CLDMs) further advance this framework by integrating user-defined image conditions through specialized conditioning modules, enabling precise control in conditional image synthesis~\cite{zhang2023adding, li2025controlnet}. By performing the denoising process in latent space, CLDMs strive to balance computational efficiency with the preservation of fine details.

Recently, Conditional Latent Diffusion Models (CLDMs) have seen a surge in adoption for image restoration (IR) tasks~\cite{wang2024exploiting, lin2024diffbir, chen2024hierarchical, yang2023pasd, yu2024scaling, xia2023diffir, ai2024multimodal, zhang2024diffbody}. However, despite these advancements, the effectiveness of CLDMs for IR tasks remains unclear. While these models leverage architectures originally developed for conditional image generation, fundamental differences exist between generative tasks and low-level IR tasks. 
Generative tasks focus on producing visually plausible outputs conditioned on high-level semantics, whereas IR demands the faithful restoration of perceptual details which often requires precise modeling of low-level representations. This raises a compelling question: \textbf{\emph{Are Conditional Latent Diffusion Models truly effective for image restoration?}}

% Our findings indicate that current Conditional Latent Diffusion Model (CLDM)-based IR solutions adopt architectures originally designed for conditional image synthesis. While this leverages the pretrained generative capabilities of CLDMs, our ablation experiments reveal that certain architectural components do not enhance restoration performance and may introduce issues such as instability and increased inference time. This suggests a misalignment between the existing CLDM architectures and the objectives of IR tasks. Unlike conditional image synthesis, which guides outputs toward conditions at a semantic level without constraining perceptual details, IR aims to guide outputs away from degraded conditions and models the relationship from condition to target using low-level perceptual features. 

To investigate this question, we conduct comprehensive experiments comparing state-of-the-art CLDM-based image restoration models with traditional deep learning approaches across various tasks. Interestingly, our findings reveal that, despite their scalability advantages, CLDM-based models often fall short in preserving fine-grained details and achieving good distortion metrics. For samples with low degradation levels, where traditional models effectively invert the degradation, CLDM-based approaches still struggle to deliver satisfactory results. Furthermore, CLDMs often introduces semantic deviations in the restored images, which is particularly problematic for restoration tasks requiring precise fidelity. To evaluate the semantic deviation issue, we introduce "Alignment" as a new evaluation aspect, which also facilitates the assessment of real-world blind image restoration.

% We also find that traditional evaluation methods, particularly those based on the distortion-perception tradeoff, face challenges in effectively assessing the performance of real-world blind image restoration because of the complexity of real-world degradation process. This evaluation method fail to capture the essential requirement of maintaining consistency between restored images and their degraded inputs. This work introduces alignment as a new evaluation aspect and employs preliminary methods for assessment.

Furthermore, we perform an in-depth analysis of the CLDM architecture to evaluate how its design elements influence restoration performance. Our empirical experiments reveal that certain architectural components like multi-timesteps and latent space transformation do not enhance restoration performance much but introduce issues such as instability and increased inference time.

Our contributions are as follows:

\begin{itemize}
    \item To the best of our knowledge, this is the first work to challenge the effectiveness of the booming CLDMs for the image restoration tasks. 
    
    \item To validate our claims, we compares CLDM-based models with traditional image restoration approaches, we demonstrate that CLDM-based IR solutions exhibit issues like high distortion, semantic deviation, misalignment with resources utilization and model performance. 

    % \item To evaluate the issue of semantic deviation and overcome the limitations of traditional distortion-perception tradeoffs in assessing real-world image restoration, we propose \emph{alignment} as a new evaluation aspect and employ fundamental methods for its assessment.
    
    \item We conduct an empirical analysis of critical design components in Conditional Latent Diffusion Models (CLDMs), including latent space representations, noise handling in the diffusion process, and multi-timestep sampling, to evaluate their impact on restoration quality. Our findings reveal that current CLDM-based solutions exhibit architectural misalignments with the objectives of image restoration tasks.
\end{itemize}

We hope this work inspires further in-depth exploration of CLDM-based IR solutions, the creation of improved evaluation metrics, and the development of innovative models that transcend current limitations and unlock the full potential of CLDMs in IR's application.

% These findings call for a reevaluation of the current research trajectory in image restoration, which heavily favors CLDMs. We advocate for model architectures and training strategies that align more closely with the unique demands of image restoration tasks. Our work aims to inspire the development of innovative models that better integrate generative strengths with the precision required for effective restoration.

\section{Preliminary: Image restoration}
% problem definition
% (related work)
% summary; pipeline

Image restoration seeks to recover the ground truth image $I_x$ from its degraded counterpart $I_y$, representing a classic inverse problem. Typically, the degradation process can be expressed as:
\begin{equation}
    I_y =  \mathcal{D} (I_x; \delta)\downarrow_\alpha + \beta \cdot N,
    \label{eq:formula1}
\end{equation}
where $\mathcal{D}$ denotes the degradation mapping function, $I_x$ is the corresponding ground truth image, and $\delta$ represents the parameters governing the degradation process. The term $\alpha$ denotes the downsampling factor, encapsulating the information loss during degradation, while $N$ represents additional noise, and $\beta$ is the signal-to-noise ratio.

The inverse of this degradation process is the restoration process, which aims to reconstruct a high-resolution (HR) approximation $\hat{I_x}$ of the ground truth HR image $I_x$ from the low-resolution (LR) image $I_y$:
\begin{equation}
    \hat{I_x} = \mathcal{F} (I_y; \theta),
\end{equation}
where $\mathcal{F}$ is the image restoration model, and $\theta$ represents its parameters. 
% The restoration process incorporates noise reduction, reversal of degradation effects, and detail generation to approximate the original image.

% The optimization objective for image restoration is defined as:
% \begin{equation}
%     \hat{\theta} = \mathop{\arg \min}_{\theta} \mathcal{L} (\hat{I_x}, I_x) + \lambda \Phi (\theta),
% \end{equation}
% where $\mathcal{L} (\hat{I_x}, I_x)$ is the loss function quantifying the discrepancy between the restored image $\hat{I_x}$ and the ground truth image $I_x$. The term $\Phi (\theta)$ is a regularization function to constrain the model's parameters, and $\lambda$ is the trade-off term balancing the reconstruction loss and the regularization term.

\section{CLDM-based IR Pipeline}

% \begin{figure*}[t]
% \vspace{-0.2cm}
% \begin{center}
% \includegraphics[width=1\textwidth]{figures/framework.png}
% \end{center}
% \vspace{-0.4cm}
% \caption{The framework of existing CLDM-based IR solutions.}
% \vspace{-0.6cm}
% \label{fig:pipeline}
% \end{figure*}

\subsection{Initial Restoration Module}
In CLDM-based image restoration, a conventional IR model is often employed as the first stage to mitigate degradations such as noise and compression artifacts in low-quality (LQ) images. The initially restored image $I_{\text{reg}}$ is computed as:
\begin{equation}
    I_{\text{reg}} = \mathcal{M}_{\phi}(I_y),
\end{equation}
where $\mathcal{M}_{\phi}$ represents the restoration module with parameters $\phi$, and $I_y$ is the LQ input image. The restoration model is typically trained using an $L_2$ pixel-wise loss, defined as:
\begin{equation}
    L_{\text{reg}} = \| I_{\text{reg}} - I_x \|_2^2,
\label{reg}
\end{equation}
where $I_x$ denotes the high-quality target image.

\subsection{Perceptual Image Compression}
To reduce computational complexity, images are projected from the high-dimensional pixel space into a lower-dimensional latent space using a perceptual compression model, comprising an encoder $\mathcal{E}$ and a decoder $\mathcal{D}$. The encoder maps an image $x \in \mathbb{R}^{H \times W \times 3}$ to a latent representation $z = \mathcal{E}(x)$, while the decoder reconstructs the image as $\tilde{x} = \mathcal{D}(z)$. This compression process, which involves downsampling, inevitably introduces information loss, quantified by:
\begin{equation}
    \mathcal{L}_{\text{info\_loss}} = \| x - \tilde{x} \|_2^2,
\label{info}
\end{equation}

\subsection{Conditioning Modules}

In CLDMs, conditioning modules guide the diffusion process using both text and image conditions, which are typically derived from the initially restored image $I_{\text{reg}}$.

\textbf{Text Conditioning}: A text encoder $\tau_{\theta}$ encodes text prompts generated from $I_{\text{reg}}$ using an image-to-text model $\mathcal{T}$:
\begin{equation}
    I_{\text{text}} = \mathcal{T}(I_{\text{reg}}), \quad c_t = \tau_{\theta}(I_{\text{text}}).
\end{equation}
% Text conditioning provides semantic-level guidance, which often does not align well with the low-level objectives of image restoration. Consequently, some works disable this component to avoid mismatched guidance~\cite{lin2024diffbir}.

\textbf{Spatial Conditioning}: The latent encoding of the initially restored image $I_{\text{reg}}$ serves as the image condition:
\begin{equation}
    c_f = \mathcal{E}(I_{\text{reg}}),
\end{equation}
where $\mathcal{E}$ is the encoder. A connector module further maps features from $c_f$ to guide the denoising network during the diffusion process. In most CLDM-based IR models, the spatial conditioning module is the primary trainable component.

\subsection{Training Process}
% The CLDM is trained to reconstruct the original image from noisy latent representations, guided by conditioning inputs. 
The training process Starts with the latent encoding of the ground truth image, $z_0 = \mathcal{E}(I_x)$, noise is progressively added to obtain $z_t$ at timestep $t$:
\begin{equation}
    z_t = \sqrt{\bar{\alpha}_t} \, z_0 + \sqrt{1 - \bar{\alpha}_t} \, \epsilon,
\end{equation}
where $\epsilon \sim \mathcal{N}(0, I)$ represents Gaussian noise, and $\bar{\alpha}_t$ denotes the cumulative product of noise scheduling factors.

The model predicts the noise $\epsilon_{\theta}(z_t, t, c_t, c_f)$ through a denoising process. The training objective minimizes the following loss:
\begin{equation}
    \mathcal{L}_{\text{CLDM}} = \mathbb{E}_{t, z_0, \epsilon} \left[ \| \epsilon - \epsilon_{\theta}(z_t, t, c_t, c_f) \|^2 \right],
\end{equation}
where $c_t$ and $c_f$ are text and spatial conditioning inputs, respectively. To ensure the model learns denoising across different noise levels, timesteps $t$ are sampled uniformly from $\{1, \dots, T\}$ during training.

\subsection{Sampling Process}
During sampling, the model iteratively denoises from a random initial latent $z_T \sim \mathcal{N}(0, I)$, guided by the conditioning inputs $c_t$ and $c_f$. At each timestep $t$, the model predicts $\epsilon_{\theta}(z_t, t, c_t, c_f)$ and updates the latent variable as follows:
\begin{equation}
    z_{t-1} = \frac{1}{\sqrt{\alpha_t}} \left( z_t - \frac{1 - \alpha_t}{\sqrt{1 - \bar{\alpha}_t}} \, \epsilon_{\theta}(z_t, t, c_t, c_f) \right) + \sigma_t \, \epsilon',
\end{equation}
where $\alpha_t$ is the noise schedule factor, $\sigma_t$ is the standard deviation of the added noise, and $\epsilon' \sim \mathcal{N}(0, I)$ represents Gaussian noise. After iteratively denoising to $z_0$, the final restored image is obtained by decoding the latent variable:
\begin{equation}
    \hat{I}_x = \mathcal{D}(z_0),
\end{equation}

\section{Issues in CLDM-based IR Solutions}

\textbf{High Distortion Issue}

Classic Image Restoration (CIR) tasks are traditionally defined under constrained settings with simple and well-defined degradation mappings $\mathcal{D}$, such as bicubic downsampling. These tasks are typically evaluated using distortion metrics, which quantify the deviation between the restored image and the ground truth, thereby assessing the model's ability to accurately invert the degradation process. The perception-distortion tradeoff, introduced by Blau~\cite{blau2018perception}, extends this evaluation by incorporating perception as a metric, emphasizing the naturalness and realism of the restored image.

\begin{figure*}
    \centering
    \begin{subfigure}[t]{0.48\textwidth}
        \centering     
        \includegraphics[width=\textwidth]{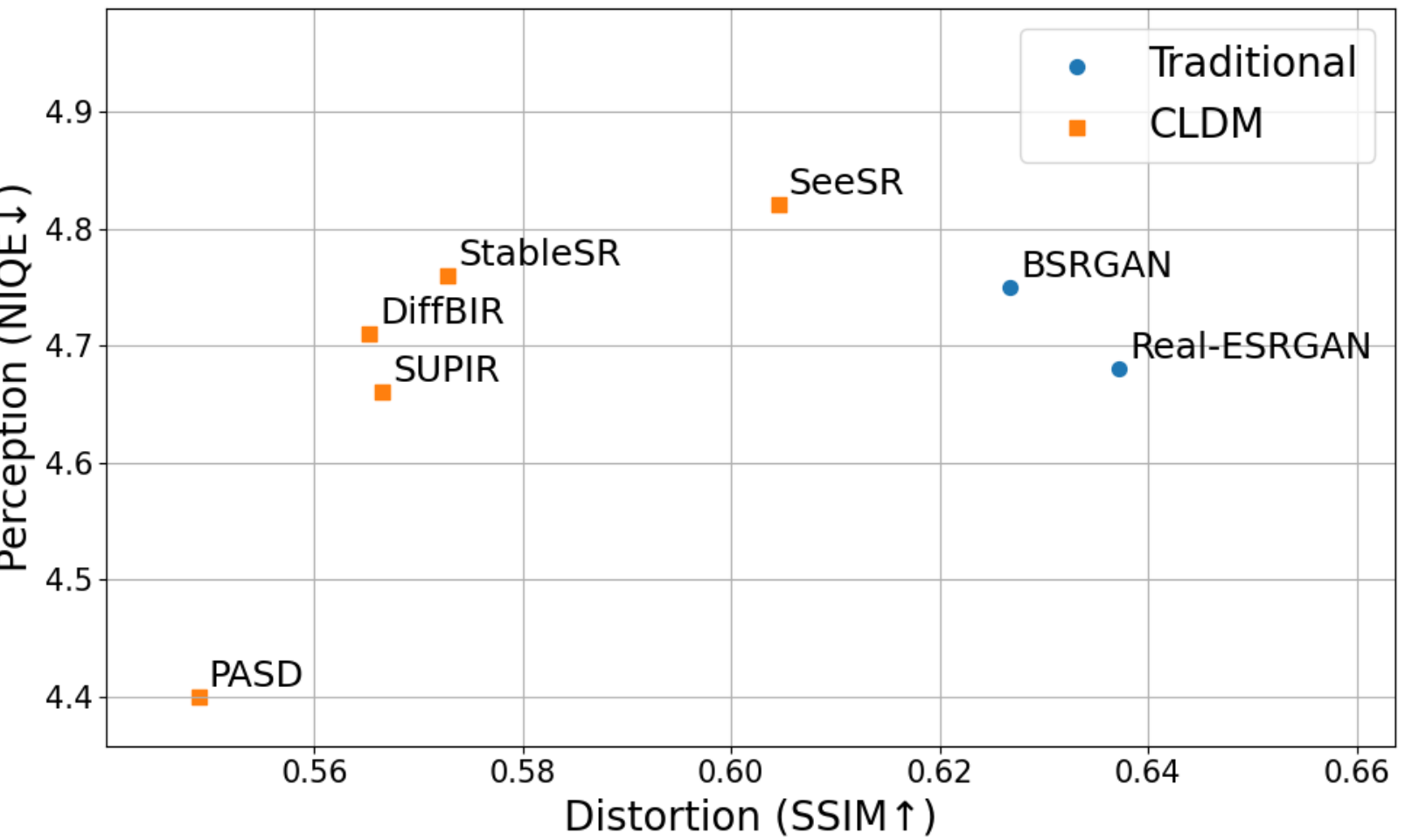}
        \caption{Super-resolution performance in DIV2K\cite{Timofte_2017_CVPR_Workshops}. }
        \label{fig:Super-resolution}
    \end{subfigure}
    \hfill
    \begin{subfigure}[t]{0.48\textwidth}
        \centering     
        \includegraphics[width=\textwidth]{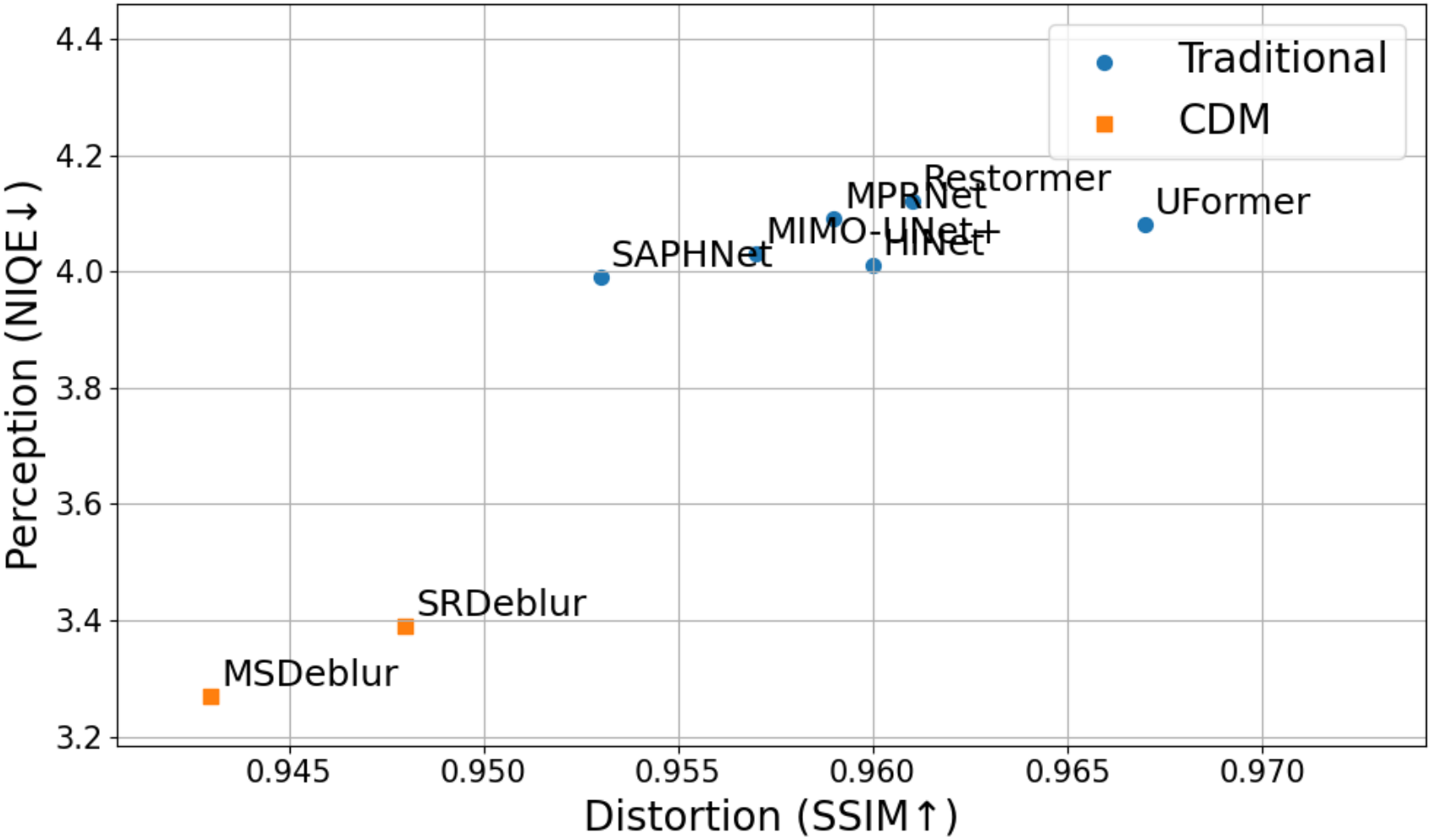}
        \caption{Deblur performance in GOPRO\cite{Nah_2017_CVPR}}
        \label{fig:Deblur}
    \end{subfigure}
    \caption{Perception-distortion tradeoff on CIR tasks.}
    \label{fig:distortion_perception_CIR}
\end{figure*}

We compare traditional models and CLDM-based models in two common CIR tasks: super-resolution and deblurring (Fig.~\ref{fig:distortion_perception_CIR}). While CLDMs demonstrate advantages in perceptual quality, they typically exhibit higher distortion. However, tn CIR, distortion serves as the primary evaluation criterion, reflecting a model's ability to accurately model the restoration process under constrained settings, such as specific tasks and datasets.

\begin{figure*}
    \centering
    \begin{subfigure}[t]{0.48\textwidth}
        \centering     
        \includegraphics[width=\textwidth]{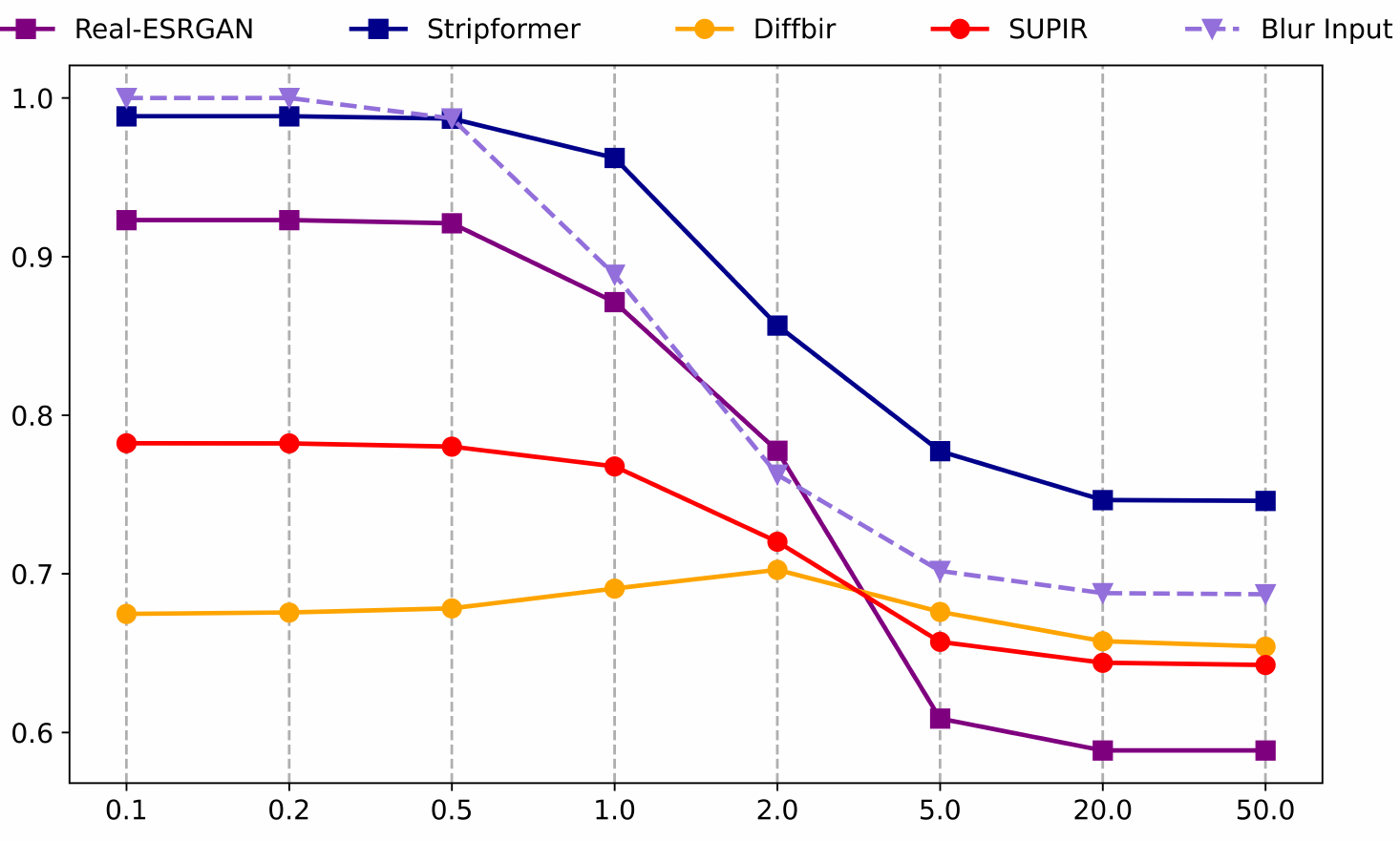}
        \caption{SSIM scores vs. Gaussian blur levels}
        \label{fig:distortion}
    \end{subfigure}
    % \hfill
    \begin{subfigure}[t]{0.48\textwidth}
        \centering     
        \includegraphics[width=\textwidth]{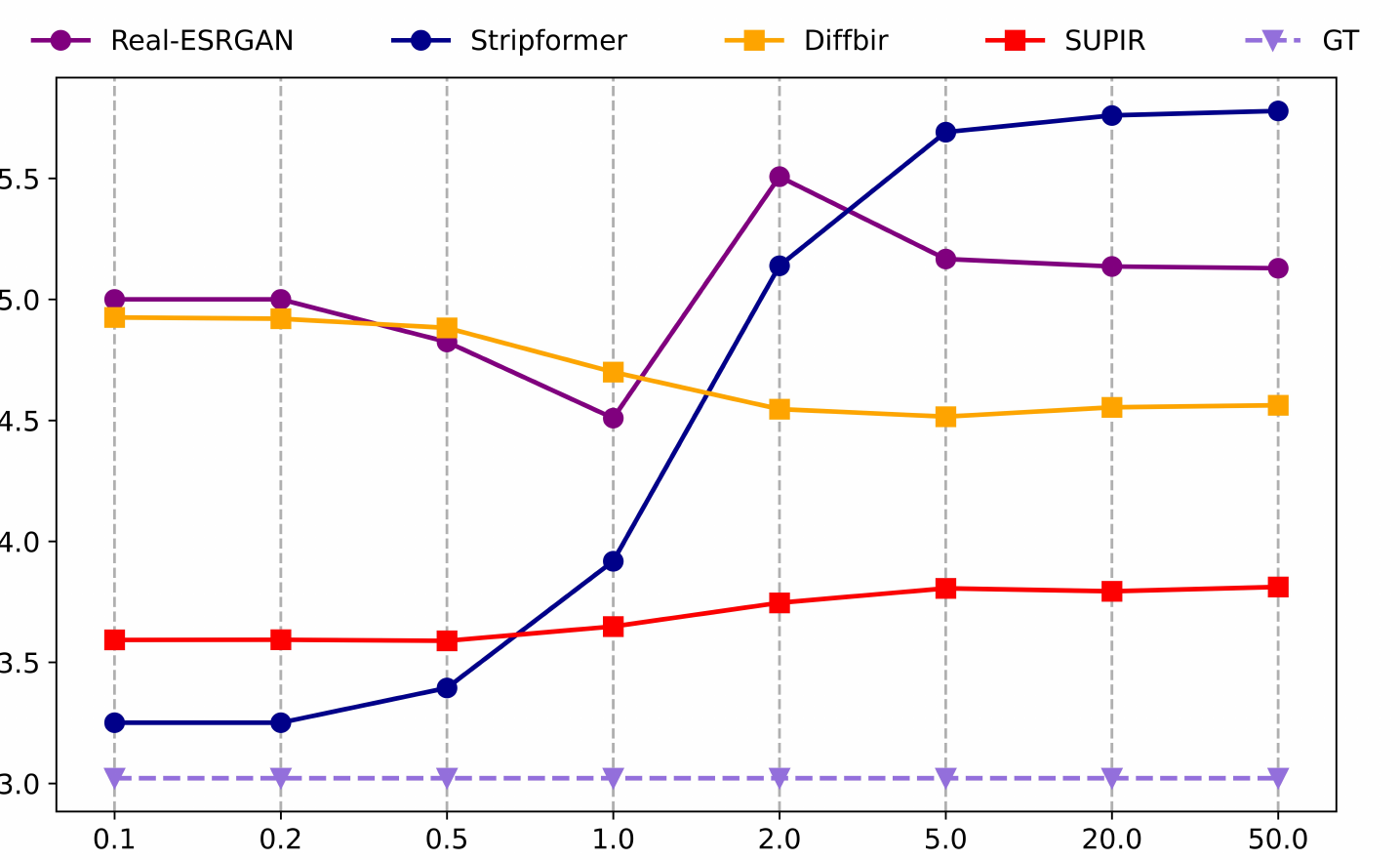}
        \caption{NIQE scores vs. Gaussian blur levels}
        \label{fig:perception}
    \end{subfigure}
    \caption{Performance comparison on varying Gaussian blur levels.}
\label{fig:different_degradation_levels_distortion_perception}
\end{figure*}

\begin{figure*}
    \centering
    \begin{subfigure}[t]{0.48\textwidth}
        \centering     
        \includegraphics[width=\textwidth]{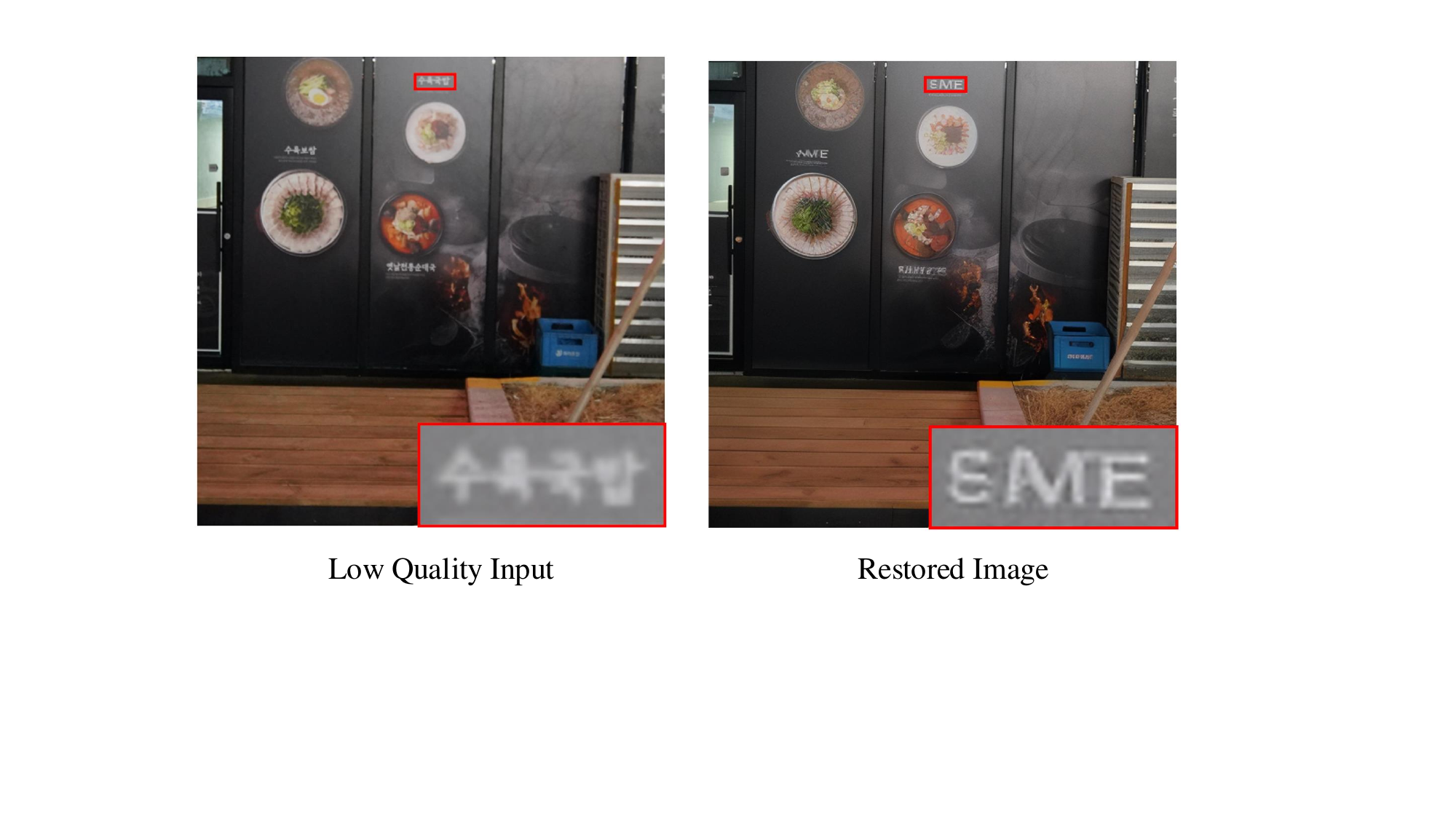}
        % \caption{Example 1}
        \label{fig:semantic_example1}
    \end{subfigure}
    % \hspace{0.1\textwidth}
    \begin{subfigure}[t]{0.48\textwidth}
        \centering     
        \includegraphics[width=\textwidth]{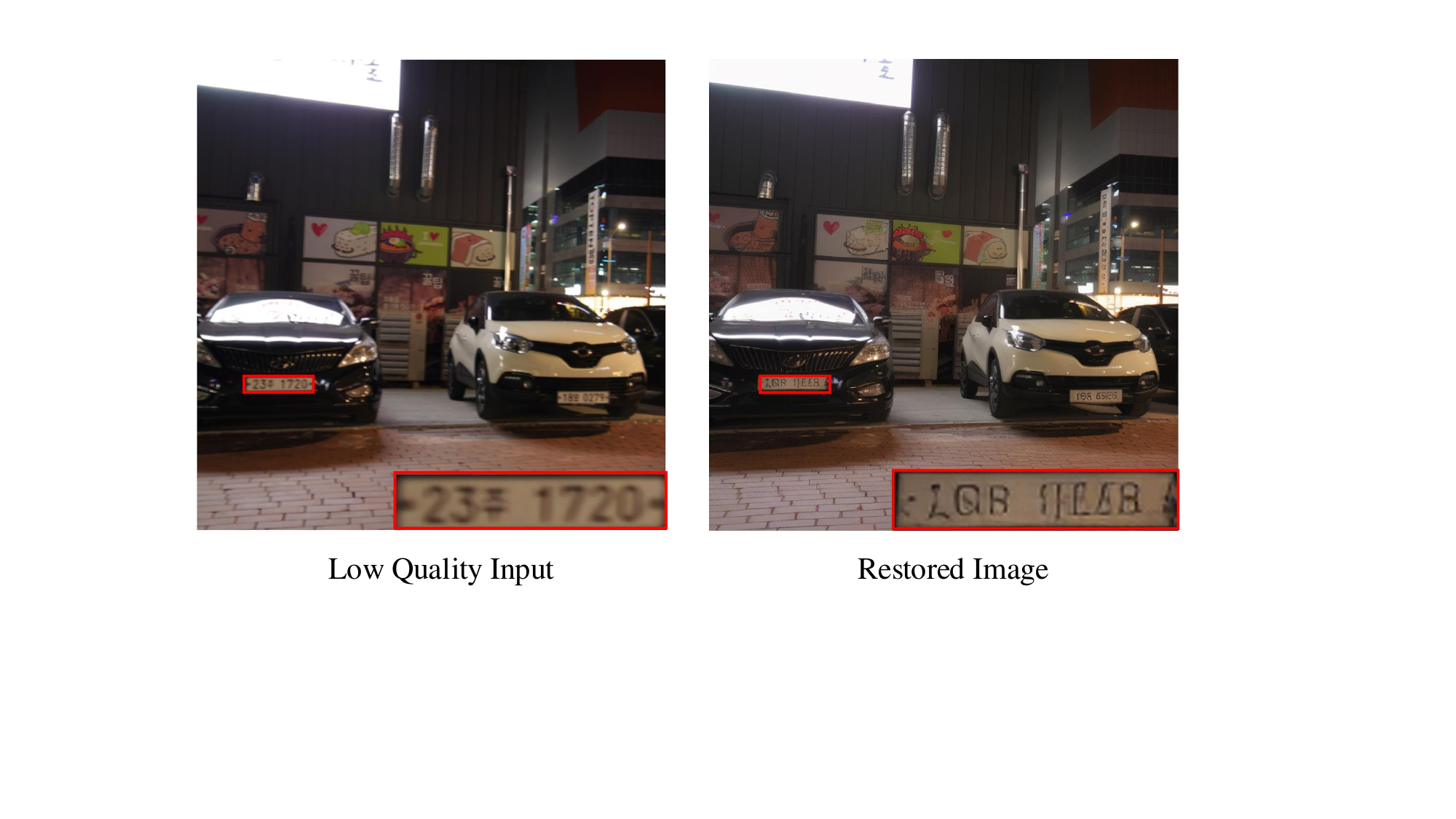}
        % \caption{Example 2}
        \label{fig:semantic_example2}
    \end{subfigure}
    \vspace{-0.5cm}
    \caption{Examples illustrating semantic deviation in CLDM outputs (DIFFBIR).}
    \label{fig:semantic_deviation}
\end{figure*}

% Traditional perception-distortion evaluations often focus on the entire validation set, overlooking the impact of varying degradation levels. We define the information retention rate $\gamma$ during the degradation process as:
Traditional perception-distortion evaluations typically treat the entire validation set uniformly, failing to account for the varying degradation levels present in individual samples. To better characterize this effect, we introduce the information retention rate during the degradation process, defined as:
\begin{equation}
\gamma = \frac{1 - \beta}{\alpha},
\end{equation}
where $\beta$ represents the degradation factor and $\alpha$ denotes the downsampling rate. 
As the degradation level increases, $\gamma$ decreases, making it more challenging to accurately estimate the inverse restoration process. 

To explore these dynamics, we evaluate four models on the DIV2K validation dataset~\cite{Timofte_2017_CVPR_Workshops}: two traditional methods (Real-ESRGAN~\cite{wang2021realesrgan} and Stripformer~\cite{tsai2022stripformer}) and two state-of-the-art CLDM-based methods (DIFFBIR~\cite{lin2024diffbir} and SUPIR~\cite{yu2024scaling}). All models are trained using similar degradation simulation pipelines. The Structural Similarity Index (SSIM) is employed to quantify distortion, while the Natural Image Quality Evaluator (NIQE)~\cite{mittal2012making} is used to assess perceptual quality.

% We apply Gaussian blur with varying standard deviations to the ground truth images to simulate different degradation levels. As illustrated in Fig.~\ref{fig:different_degradation_levels_distortion_perception}, traditional models excel at low degradation levels, producing outputs that score highly in both distortion and perception metrics. However, as the level of degradation intensifies, their performance drops dramatically. In contrast, CLDM-based models maintain consistently high perceptual quality across a wide range of degradation levels, yet they exhibit persistently high distortion—even when the input images are of good quality. 

% Conversely, when the degradation level is low, the degraded images naturally resemble their high-quality ground truth counterparts and the restoration process has a small upsampling factor, making it relatively easy to achieve good performance in both distortion and perception.

% At severe degradation levels, traditional models often produce over-smoothed outputs with noticeable artifacts. Meanwhile, CLDM-based models, despite suffering from high distortion, can still generate plausible samples. As described in Equation~\eqref{eq:formula1}, the restoration process involves more than just reversing a given degradation pattern; it also requires generating fine details for the upsampling stage. Ultimately, these results highlight a trade-off: diffusion-based methods are particularly adept at generating realistic content, while traditional approaches are better at detecting and mitigating the effects of degradation.

As illustrated in Fig.~\ref{fig:different_degradation_levels_distortion_perception}, traditional models excel at low degradation levels, achieving high scores in both distortion and perception metrics. In these scenarios, the degraded images closely resemble their high-quality ground truths, and the upsampling factor is small, making restoration less challenging. Under such conditions, it is naturally expected that the outputs should not exhibit noticeable distortion, and indeed, traditional methods outperform CLDM-based models.

However, as the degradation level intensifies, restoration becomes significantly more difficult. Traditional models, which rely heavily on explicit degradation detection and compensation, start to produce over-smoothed outputs with visible artifacts. In contrast, CLDM-based models maintain high perceptual quality even under severe degradation. This indicates that diffusion-based methods, while adept at generating realistic details, struggle to ensure fidelity to the original degraded content.

As described in Equation~\eqref{eq:formula1}, the restoration process involves not only reversing the degradation pattern but also synthesizing fine details required for upsampling. Ultimately, these findings highlight a fundamental trade-off: diffusion-based methods are exceptionally skilled at producing plausible, visually appealing outputs, whereas traditional models are more effective at mitigating degradation and preserving the original image structure, particularly when the degradation level is low.

\noindent
\textbf{Semantic Deviation Issue}

We find that Conditional Latent Diffusion Models (CLDMs) often alter semantic details during the restoration process, leading to deviations from the original input semantics (see Fig.~\ref{fig:semantic_deviation}). Traditional models rely on low-level features such as frequency components and gradients, which inherently lack the capacity to inject new semantic information. In contrast, CLDMs appear to model the restoration process at a semantic level like their generative modeling. As a result, they frequently introduce semantic inconsistencies or "hallucinations," distorting the intended meaning of the original images.

There are currently no established metrics for evaluating semantic deviation in image restoration tasks. To address this limitation, we propose \textit{Alignment} as a new evaluation criterion to measure the semantic consistency between restored images and their degraded inputs at a fine-grained level. Formally, we define Alignment as:
\begin{equation}
    \text{Alignment} = \mathcal{S}(I_y, \hat{I}_x),
\end{equation}
where \(\mathcal{S}(\cdot,\cdot)\) denotes a semantic similarity measure tailored for image restoration.

Measuring Alignment in IR tasks presents notable challenges. Conventional distortion metrics capture only pixel-level discrepancies, failing to reflect semantic consistency. While semantic encoders such as CLIP and DINOv2 capture high-level concepts, they often lack the fine-grained precision required for detailed semantic comparisons. Moreover, comparing images subjected to different degradation levels introduces additional complexity. Despite these difficulties, we propose a practical, if approximate, approach.

For low-degradation images, we assume that the restored image \(\hat{I}_x\) maintains strong semantic consistency with the ground truth (GT) image \(I_x\). Under this assumption, we approximate Alignment as:
\begin{equation}
    \text{Alignment} \approx \mathcal{S}(I_x, \hat{I}_x).
\end{equation}

For high-degradation images, direct comparison with the GT image may not accurately reflect semantic alignment due to significant semantic shifts. Instead, we apply the same degradation function \(\mathcal{D}\) used to produce \(I_y\) to the restored image \(\hat{I}_x\), yielding \(\mathcal{D}(\hat{I}_x)\):
\begin{equation}
    \text{Alignment} \approx \mathcal{S}\bigl(\mathcal{D}(\hat{I}_x), I_y\bigr).
\end{equation}
This approach enables a comparison under equivalent degradation conditions, though it becomes infeasible if \(\mathcal{D}\) is unknown.

\begin{table}
    \centering
    \begin{subtable}[t]{0.48\textwidth}
        \centering
        % \label{tab:low_degradation_gt}
        \scalebox{1}{
            \begin{tabular}{llcc}
                \toprule
                \textbf{Task} & \textbf{Model} & \textbf{DINOv2$\downarrow$} & \textbf{SSIM$\uparrow$} \\
                \midrule
                \multirow{5}{*}{SR2} 
                & Stripformer  & \bfseries{0.1740}                  & \bfseries{0.9149}       \\
                & Real-ESRGAN  & -                     & -                     \\
                & DiffBIR      & 0.5846                  & 0.6987                  \\
                & SUPIR        & 0.3839       & 0.7558                  \\
                \midrule
                \multirow{5}{*}{Blur1} 
                & Stripformer  & \bfseries{0.0953}       & \bfseries{0.9622}       \\
                & Real-ESRGAN  & 0.3166                  & 0.8713                  \\
                & DiffBIR      & 0.5971                  & 0.6907                  \\
                & SUPIR        & 0.3309                  & 0.7677                  \\
                \bottomrule
            \end{tabular}
        }
        \vspace{1mm} % Adjust spacing
        \caption{Compared with \textbf{Ground Truth}}
    \end{subtable}
    \hfill
    \begin{subtable}[t]{0.48\textwidth}
        \centering
        % \label{tab:high_degradation_lq}
        % \vspace{-8mm} % Adjust spacing to align bottom
        \scalebox{1}{
            \begin{tabular}{llcc}
                \toprule
                \textbf{Task} & \textbf{Model} & \textbf{DINOv2$\downarrow$} & \textbf{SSIM$\uparrow$} \\
                \midrule
                \multirow{5}{*}{SR4} 
                & Stripformer  & \bfseries{0.5319}       & \bfseries{0.9868}       \\
                & Real-ESRGAN  & 0.6277                  & 0.9671                  \\
                & DiffBIR      & 0.5911                  & 0.9470                  \\
                & SUPIR        & 0.5613                  & 0.9553                  \\
                \midrule
                \multirow{5}{*}{Blur20} 
                & Stripformer  & 0.6088       & \bfseries{0.9761}       \\
                & Real-ESRGAN  & 0.6153                  & 0.9393                  \\
                & DiffBIR      & 0.6696                  & 0.9536                  \\
                & SUPIR        & \bfseries{0.5476}                  & 0.9520                  \\
                \bottomrule
            \end{tabular}
        }
        \vspace{1mm} % Adjust spacing
        \caption{Compared with \textbf{Low Quality Input}}
    \end{subtable}
    \vspace{-2mm}
    \caption{Alignment evaluation. Highlight the best score in \textbf{bold}.}
    \label{tab:alignment_evaluation}
\end{table}

We employed these methods to assess semantic deviations in four tasks: two high-degradation tasks (super-resolution \(\times 4\), blur \(\sigma = 20\)) and two low-degradation tasks (super-resolution \(\times 2\), blur \(\sigma = 1\)). As shown in Table~\ref{tab:alignment_evaluation}, CLDM-based models fail to maintain semantic consistency with the input, even for low-degradation tasks. While our simple approach leverages existing metrics and is therefore limited, our findings highlight the pressing need for more sophisticated Alignment evaluation methods in image restoration research.

Introducing alignment as a new evaluation aspect helps address the challenges of assessing performance in real-world Blind Image Restoration (BIR). While Classic Image Restoration (CIR) operates within a constrained domain, BIR tackles the restoration of real-world degraded images without constraints on the degradation mapping functions $\mathcal{D}$. BIR models are designed to handle complex, multiple, and even mixed types of degradations, demonstrating strong generalization capabilities.

Traditionally, both CIR and BIR have relied on distortion metrics for evaluation. In CIR tasks, the distribution of ground truth images and the degradation processes conform to specific assumptions, making the degradation mapping functions relatively simple and the corresponding inverse process approximations easier to obtain. In contrast, the real-world degradation processes in BIR are more complex, leading to diverse inverse mappings and a larger valid solution domain for the restoration process, where the ground truth image is merely one plausible sample. Consequently, distortion metrics alone are insufficient to comprehensively evaluate restoration quality in BIR tasks. Furthermore, real-world degraded images often lack corresponding ground truth references, adding to the challenge of accurate assessment.

Due to these limitations, some CLDM-based approaches focus on perception evaluation. However, perception is typically assessed using non-reference image quality assessment metrics, which fail to capture the consistency requirements essential in image restoration tasks. In this context, we propose using a combination of alignment and perception, specifically an "alignment-perception tradeoff," as a more effective evaluation framework for BIR tasks.

\noindent
\textbf{Performance vs. Resource Utilization}

%% same column

% \begin{figure}
%     \centering
%     \begin{subfigure}[t]{0.48\textwidth}
%         \centering     
%         \includegraphics[width=\textwidth]{figures/model_parameters_comparison.pdf}
%         \caption{Model parameters vs. performance}
%         \label{fig:model_parameters_performance}
%     \end{subfigure}
%     \hfill
%     \begin{subfigure}[t]{0.48\textwidth}
%         \centering     
%         \includegraphics[width=\textwidth]{figures/latency_comparison.pdf}
%         \caption{Latency vs. performance}
%         \label{fig:latency_performance}
%     \end{subfigure}
%     \caption{Comparison of performance relative to model parameters and latency.}
%     \label{fig:scale_performance_div2k}
% \end{figure}

%% same row intext.

% \begin{figure}[ht]
%     \centering
%     \begin{subfigure}[t]{0.47\linewidth}
%         \centering     
%         \includegraphics[width=\linewidth]{figures/model_parameters_comparison.pdf}
%         \caption{Model parameters vs. performance}
%         \label{fig:model_parameters_performance}
%     \end{subfigure}
%     \hfill
%     \begin{subfigure}[t]{0.47\linewidth}
%         \centering     
%         \includegraphics[width=\linewidth]{figures/latency_comparison.pdf}
%         \caption{Latency vs. performance}
%         \label{fig:latency_performance}
%     \end{subfigure}
%     \caption{Comparison of performance relative to model parameters and latency.}
%     \label{fig:scale_performance_div2k}
% \end{figure}

%% same row: not in-text
\begin{figure*}
    \centering
    \begin{subfigure}[t]{0.48\textwidth}
        \centering     
        \includegraphics[width=\textwidth]{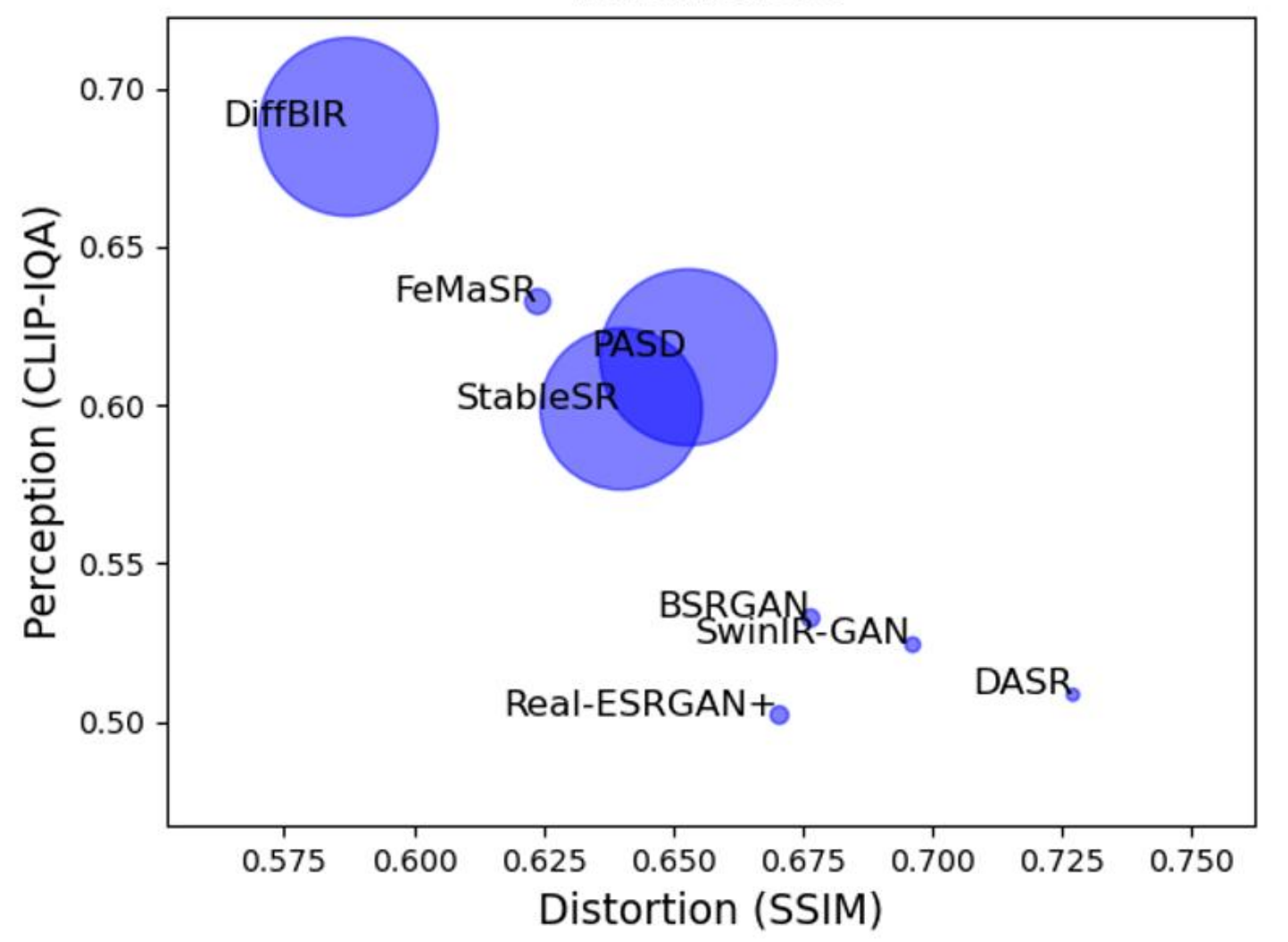}
        \caption{Model parameters vs. performance}
        \label{fig:model_parameters_performance}
    \end{subfigure}
    \hfill
    \begin{subfigure}[t]{0.48\textwidth}
        \centering     
        \includegraphics[width=\textwidth]{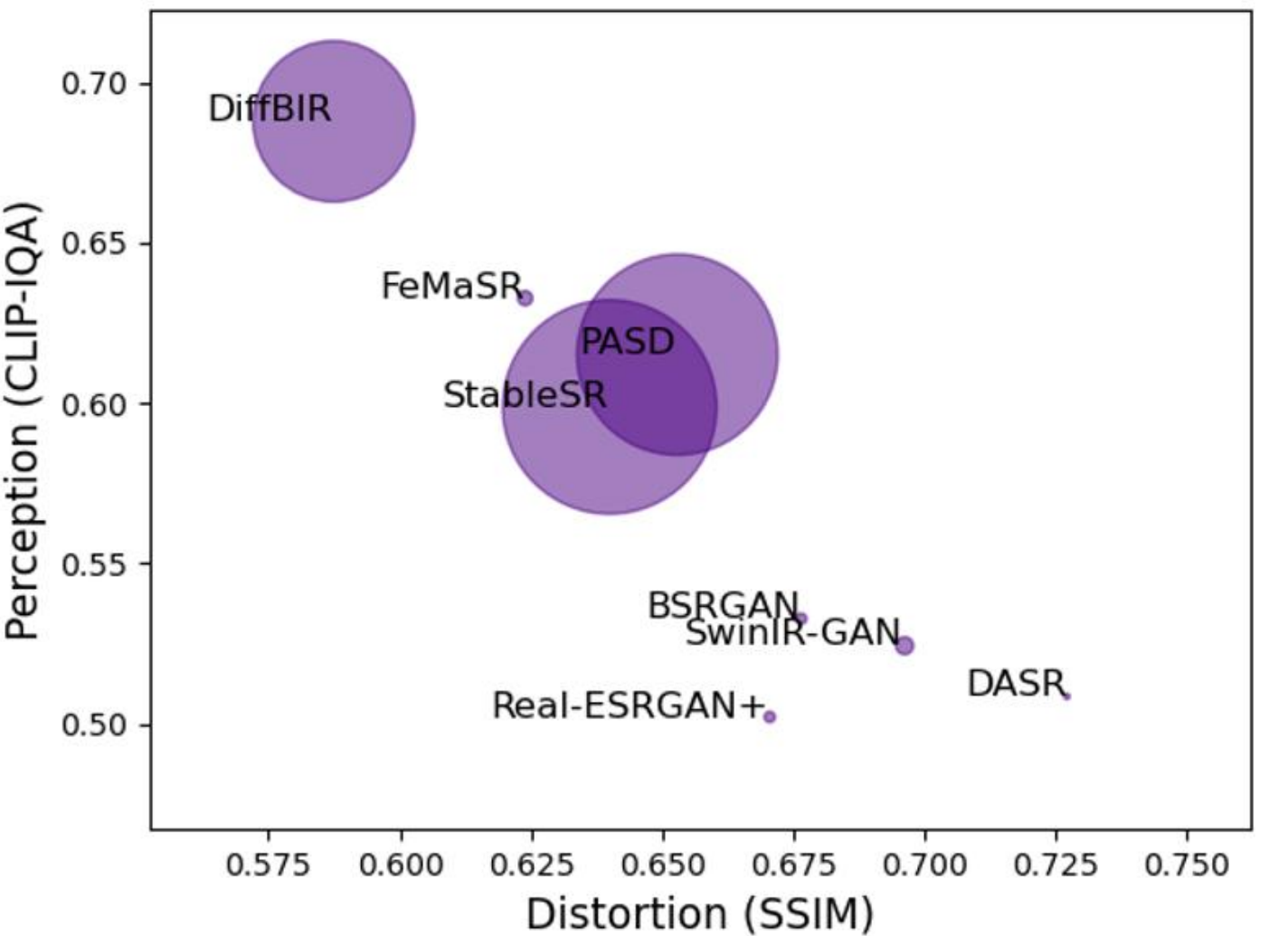}
        \caption{Latency vs. performance}
        \label{fig:latency_performance}
    \end{subfigure}
    \caption{Comparison of performance relative to model parameters and latency.}
    \label{fig:scale_performance_div2k}
\end{figure*}

% CLDM-based models exhibit significant scale advantages in terms of model size, as well as the volume of data utilized during both training and pretraining stages.For instance, models like SUPIR leverage substantial computational power and are trained on millions of images, with backbones pre-trained on datasets containing billions of images. However, despite these extensive resources, the performance improvements over traditional models remain limited and intrigures new issues such as high distortion and semantic deviations, revealing a fundamental gap between resource utilization and model performance.

% Notably, the application of CLDMs in image restoration tasks directly adopts architectural designs originally developed for conditional image synthesis. However, the intrinsic nature of these two tasks is fundamentally different. These findings suggest that the current architectural designs of CLDMs may not be well-suited to the specific demands of image restoration tasks and could even hinder their potential. Furthermore, the observed improvements may primarily stem from scaling up resources rather than an inherent correlation between the CLDM architecture and the demands of image restoration tasks.

CLDM-based models exhibit substantial advantages in terms of model scale, as well as the quantity of data employed during both training and pretraining stages. For example, certain models such as SUPIR leverage extensive computational resources and are trained on millions of images, with backbones pre-trained on datasets comprising billions of images. Despite these substantial investments, the resultant performance improvements over traditional models remain limited. Moreover, these approaches introduce new issues, including pronounced distortion and semantic deviations, thereby exposing a fundamental discrepancy between resource utilization and achievable performance.

Notably, the application of CLDMs in image restoration tasks directly inherits architectural designs originally developed for conditional image synthesis. However, the intrinsic objectives of these two tasks differ fundamentally. These observations suggest that current CLDM architectures may not be intrinsically aligned with the specific requirements of image restoration and could potentially constrain their capabilities. Furthermore, the observed improvements appear to be largely driven by increased resource allocation rather than the architectural suitability of CLDMs for image restoration tasks.

The related results are shown in Fig.~\ref{fig:scale_performance_div2k}.

\section{More Analysis on CLDMs}

% \textbf{Effectiveness of Latent Space for IR}

% Conditional Latent Diffusion Models (CLDMs) perform restoration by gradual denoising in the latent space. However, due to the downsampling in the encoding process, there is an inherent distortion limit, which can be quantified by the perceptual image compression loss $\mathcal{L}_{\text{info\_loss}}$ (see Eq.~\ref{info}). Experiments also show that the output of CLDMs cannot surpass the limit imposed by perceptual information compression (see Table~\ref{tab:quality_metrics}).

% \input{tables/quality_metrics_table}

% We also find that high-frequency details are often lost during encoding, and some semantic information is degraded during the encoding process (see Figure~\ref{fig:latent_space_effect}).

\begin{figure*}
    \centering
    \begin{subfigure}[t]{0.48\textwidth}
        \centering     
        \includegraphics[width=\textwidth]{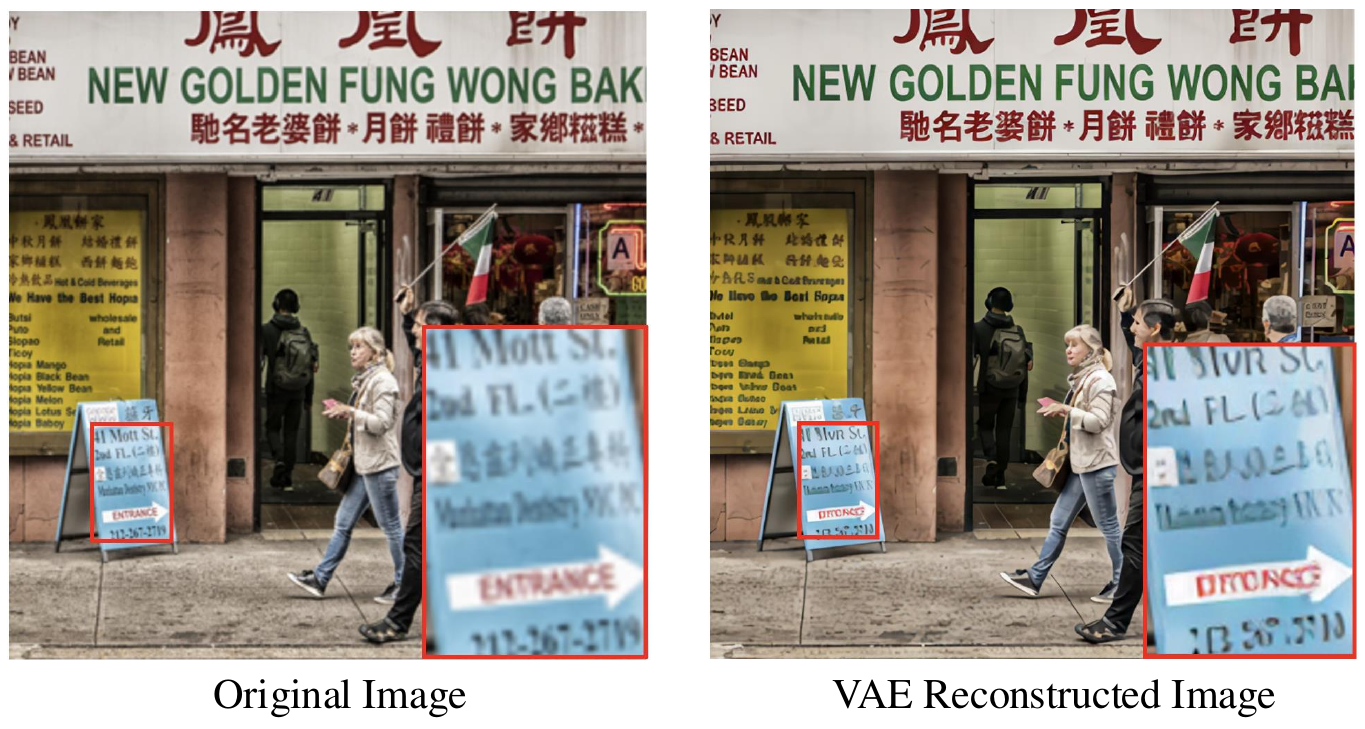}
        % \caption{Example 1}
        \label{fig:latent_space_effect1}
    \end{subfigure}
    % \hfill
    \begin{subfigure}[t]{0.48\textwidth}
        \centering     
        \includegraphics[width=\textwidth]{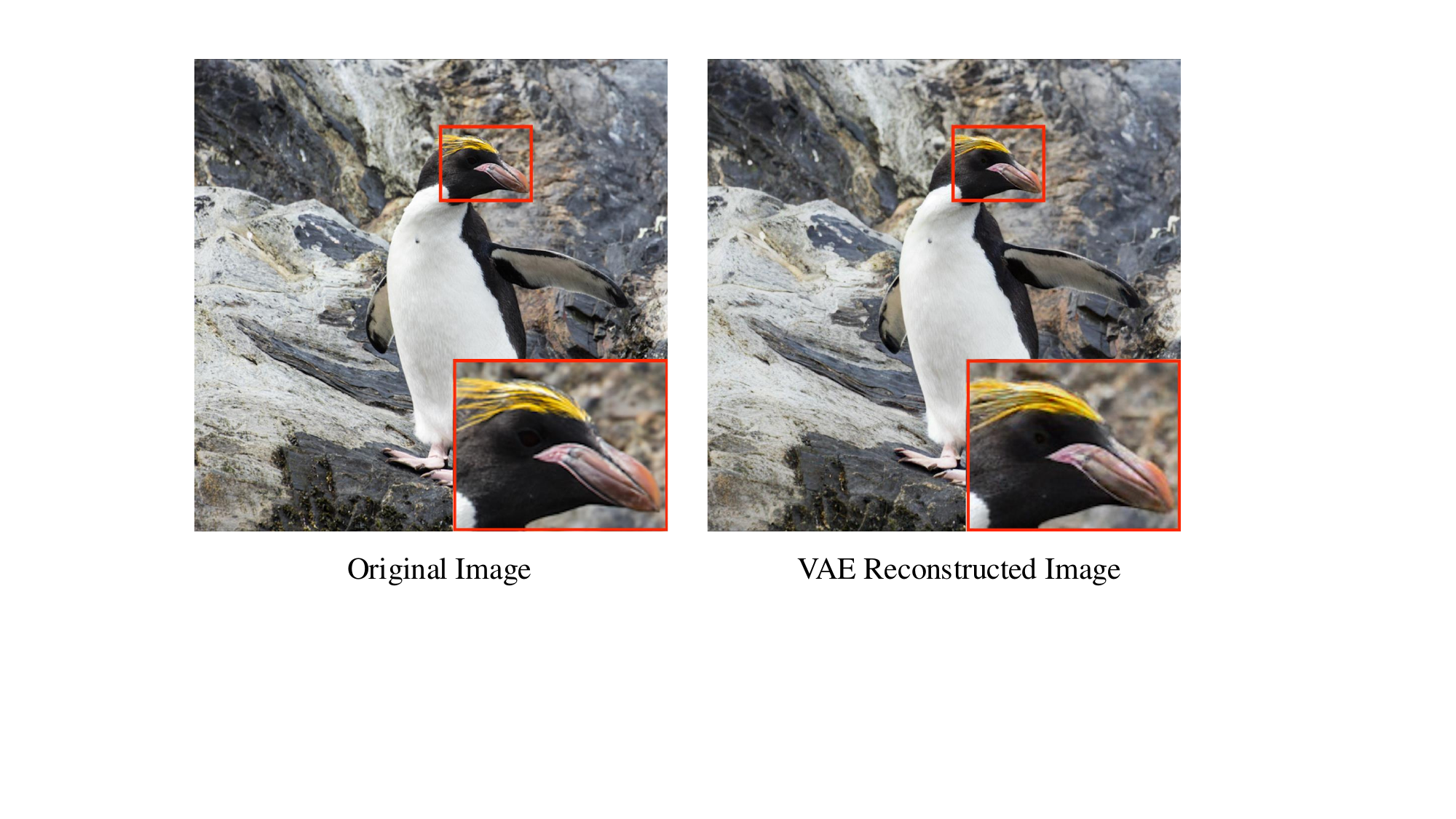}
        % \caption{Example 2}
        \label{fig:latent_space_effect2}
    \end{subfigure}
    \vspace{-0.5cm}
    \caption{Impact of latent space encoding on image details.}
    \label{fig:latent_space_effect}
\end{figure*}

% Furthermore, the diffusion loss in CLDMs is calculated in the latent space rather than the image (pixel) space. This introduces a mismatch between the training objective and the ultimate goal of IR tasks, which is to minimize distortion in the pixel space. As a result, the model may not effectively optimize for the desired restoration quality, hindering its ability to recover fine details and achieve high-fidelity reconstructions.

\noindent
\textbf{Effectiveness of Latent Space for Image Restoration}

Conditional Latent Diffusion Models (CLDMs) restore images through gradual denoising in the latent space. However, the transformation into the latent space introduces downsampling and compression, which can compromise critical perceptual details essential for accurate restoration. While this limitation does not significantly impact generative tasks, as they are not constrained by the precise perceptual fidelity of details, it significantly affects image restoration performance, which heavily relies on low-level perceptual information.

The downsampling inherent in the encoding process imposes a fundamental distortion limit, quantified by the perceptual image compression loss $\mathcal{L}_{\text{info\_loss}}$ (Eq.~\ref{info}). Experimental results further confirm that the outputs of CLDMs cannot exceed the quality constraints imposed by this compression (Table~\ref{tab:quality_metrics}).

% \begin{table}
%     \centering
%     \setlength{\tabcolsep}{2.7pt} % Adjust the column spacing
%     \scalebox{1}{
%         \begin{tabular}{lcccc}
%             \toprule
%             \textbf{   } & \textbf{PSNR$\uparrow$} & \textbf{SSIM$\uparrow$} & \textbf{LPIPS$\downarrow$} & \textbf{NIQE$\downarrow$}\\
%             \midrule
%             GT            & -      & -       & -       & 3.02\\
%             DIFFBIR-VAE   & 27.72  & 0.7894  & 0.0580  & 3.28\\
%             DIFFBIR       & 25.56  & 0.7025  & 0.1421  & 4.52\\
%             SUPIR-VAE     & 28.68  & 0.8184  & 0.0664  & 3.36\\
%             SUPIR         & 27.27  & 0.7828  & 0.0978  & 3.59\\
%             \bottomrule
%         \end{tabular}
%     }
%     \vspace{1mm} % Adjust spacing if needed
%     \caption{Image Quality Metrics Comparison}
%     \label{tab:quality_metrics}
% \end{table}

\begin{table}
    \centering
    \setlength{\tabcolsep}{2.7pt} % Adjust the column spacing
    \scalebox{1}{
        \begin{tabular}{lcccc}
            \toprule
            \textbf{ } & \multicolumn{2}{c}{\textbf{Distortion}} & \multicolumn{2}{c}{\textbf{Perceptual}} \\
            \cmidrule(lr){2-3} \cmidrule(lr){4-5}
            \textbf{ } & PSNR$\uparrow$ & SSIM$\uparrow$ & LPIPS$\downarrow$ & NIQE$\downarrow$\\
            \cmidrule(lr){1-5}
            GT          & -      & -       & -       & 3.02\\
            DIFFBIR-VAE & 27.72  & 0.7894  & 0.0580  & 3.28\\
            DIFFBIR     & 25.56  & 0.7025  & 0.1421  & 4.52\\
            SUPIR-VAE   & 28.68  & 0.8184  & 0.0664  & 3.36\\
            SUPIR       & 27.27  & 0.7828  & 0.0978  & 3.59\\
            \bottomrule
        \end{tabular}
    }
    \vspace{1mm} % Adjust spacing if needed
    \caption{Image Quality Metrics Comparison}
    \label{tab:quality_metrics}
\end{table}

            % GT            & -      & -       & -       & 3.0215 \\
            % DIFFBIR-VAE   & 27.72  & 0.7894  & 0.0580  & 3.2851 \\
            % DIFFBIR       & 25.56  & 0.7025  & 0.1421  & 4.5154 \\
            % SUPIR-VAE     & 28.68  & 0.8184  & 0.0664  & 3.3585 \\
            % SUPIR         & 27.27  & 0.7828  & 0.0978  & 3.5891 \\

Additionally, we observe that high-frequency details are often lost during the encoding process, and some semantic information is degraded (Fig.~\ref{fig:latent_space_effect}). 

Furthermore, the diffusion loss in CLDMs is calculated within the latent space, rather than the pixel space. This creates a mismatch between the training objective and the ultimate goal of image restoration tasks. 

\noindent
\textbf{Impact of Noise Levels in CLDM Sampling}

% CLDMs start sampling from random Gaussian noise, introducing randomness that enhances diversity in generative tasks. However, in IR tasks, diversity is less critical. We experimented with varying noise levels by adjusting the starting timestep and observed that higher noise levels consistently increased distortion without significantly improving perceptual quality, especially for low degradation tasks (Fig. ~\ref{fig:noise_impact}).

CLDMs initiate sampling from random Gaussian noise, introducing stochasticity that enhances diversity in generative tasks. However, in image restoration (IR) tasks, diversity is less crucial, as the primary goal is to achieve faithful reconstructions. 

We investigate the effect of varying noise levels during sampling by adjusting the starting timestep. Experimental results show that higher noise levels consistently lead to increased distortion without providing notable improvements in perceptual quality, particularly for tasks involving low degradation levels (Fig.~\ref{fig:noise_impact} in Supplementary Material).

% We used the similarity relationships among the LQ input, GT, and the output to visualize all samples, finding that higher noise levels also led to more unstable outputs and greater variability for the same input, reducing model reliability (see Figure~\ref{fig:noise_distribution}).

% \begin{figure*}[!htbp]
%     \centering
%     \begin{subfigure}[t]{0.45\textwidth}
%         \centering     
%         \includegraphics[width=\textwidth]{figures/example.pdf}
%         \caption{Noise level = 0.2}
%         \label{fig:low_noise_example}
%     \end{subfigure}
%     \hfill
%     \begin{subfigure}[t]{0.45\textwidth}
%         \centering     
%         \includegraphics[width=\textwidth]{figures/example.pdf}
%         \caption{Noise level = 1.0}
%         \label{fig:high_noise_example}
%     \end{subfigure}
%     \caption{Examples of outputs at different noise levels.}
%     \label{fig:noise_distribution}
% \end{figure*}

\noindent
\textbf{Effectiveness of Multi-Step Sampling}

CLDMs generate images through multi-step denoising, a process critical for modeling complex distributions in generative tasks. 

% However, in image restoration (IR) tasks, the primary restoration process is predominantly guided by the conditioning module.

We hypothesize that the modeling of the restoration process is primarily performed by the conditioning module, and the performance of IR is less influenced by the multi-timestep mechanism compared to generative tasks.We conducted experiments with varying numbers of sampling steps. The results indicate that increasing the number of sampling steps does not effectively reduce distortion. (Fig.~\ref{fig:sampling_steps_impact} in Supplementary Material).

% Instead, the multi-timestep mechanism primarily enhances the perceptual quality of the output images without significantly improving the accuracy of the restoration process 

To further investigate, we tested one-step prediction and found that the capability of predicting the ground truth (reflected by distortion metrics) remains nearly the same. Even when starting from pure Gaussian noise, the network achieves its optimal ability to predict the image (Fig.~\ref{fig:one_step_impact} in Supplementary Material).

% We also observe that even when starting from complete noise, the model predicts the original image relatively accurately, suggesting that the restoration process is primarily controlled by the conditioning module rather than the diffusion steps.

Additionally, we test the performance at different inference steps, further demonstrating that the multi-timestep sampling process does not improve prediction accuracy but merely enhances image quality (Fig.~\ref{fig:inference_steps_impact} in Supplementary Material).

\section{Conclusion}
%\vspace{-0.5cm}
This study questions the effectiveness of emerging CLDM-based solutions for image restoration. By comparing state-of-the-art CLDM models with traditional approaches, we revealed significant limitations of CLDMs, including high distortion, semantic deviations, and a gap between resource utilization and model performance. To address the shortcomings of traditional evaluation metrics in real-world blind image restoration, we proposed a new evaluation aspect—\textit{alignment},hoping to inspiring future more advanced evaluation methods. Our ablation studies indicate that certain design components of CLDMs do not enhance performance in image restoration, highlighting a mismatch between model architecture and task requirements. We anticipate that further analysis of CLDM applications in image restoration will be beneficial and hope that our comprehensive studies will inform and assist future research in this area.

% conclusion; limitation; future work
{
    \small
    \bibliographystyle{ieeenat_fullname}
    \bibliography{main}
}

\clearpage
\setcounter{page}{1}
\maketitlesupplementary

% Ensure you have included the necessary packages in your preamble:
% \usepackage{graphicx}
% \usepackage{caption}
% \usepackage{subcaption}
% \usepackage{booktabs}
% \usepackage{multirow}

\section{Overview}
\begin{figure}[htbp]
    \centering
    \begin{subfigure}[t]{0.48\textwidth}
        \centering     
        \includegraphics[width=\textwidth]{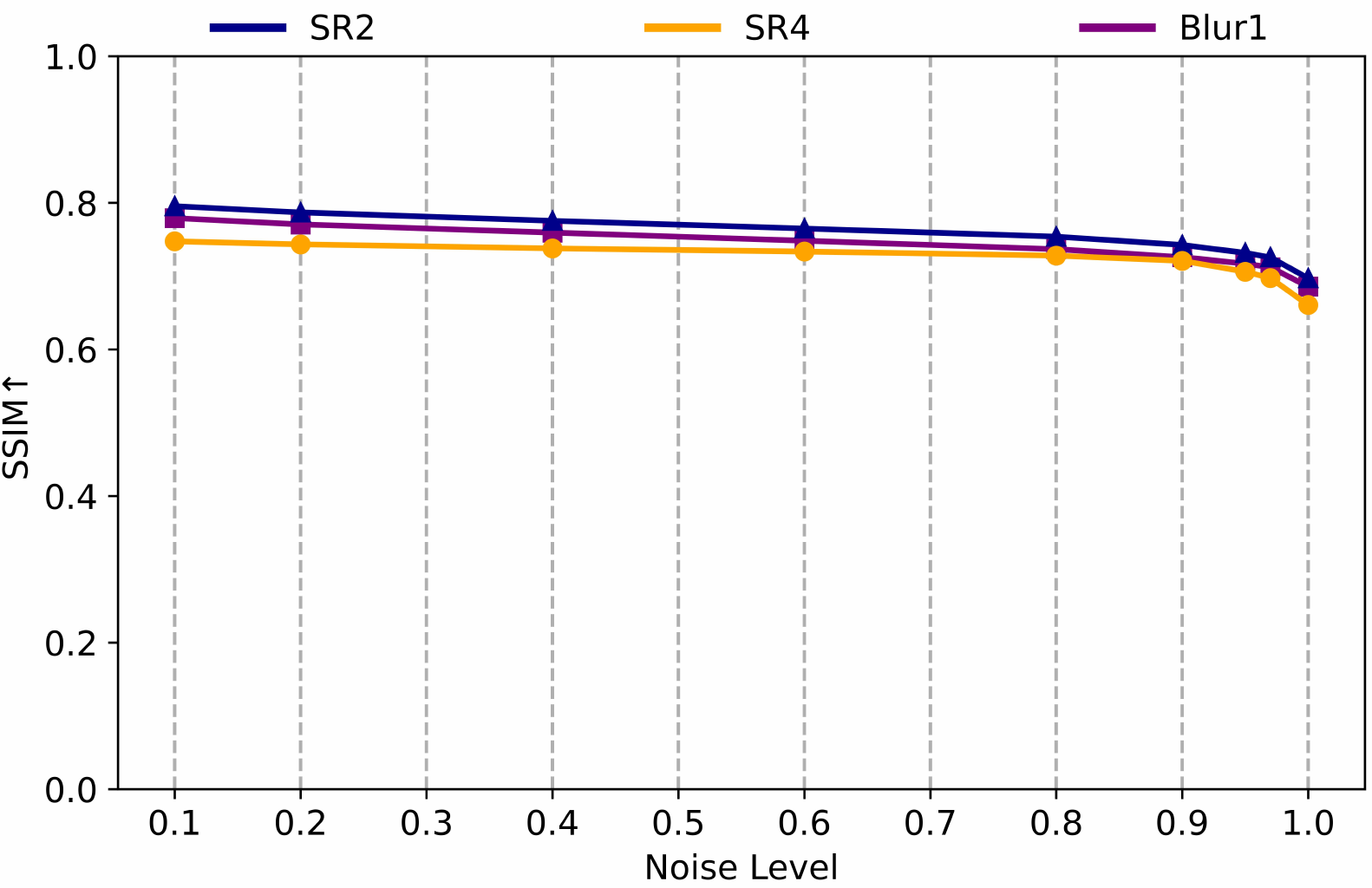}
        \caption{SSIM scores vs. noise levels}
        \label{fig:noise_distortion}
    \end{subfigure}
    \hfill
    \begin{subfigure}[t]{0.48\textwidth}
        \centering     
        \includegraphics[width=\textwidth]{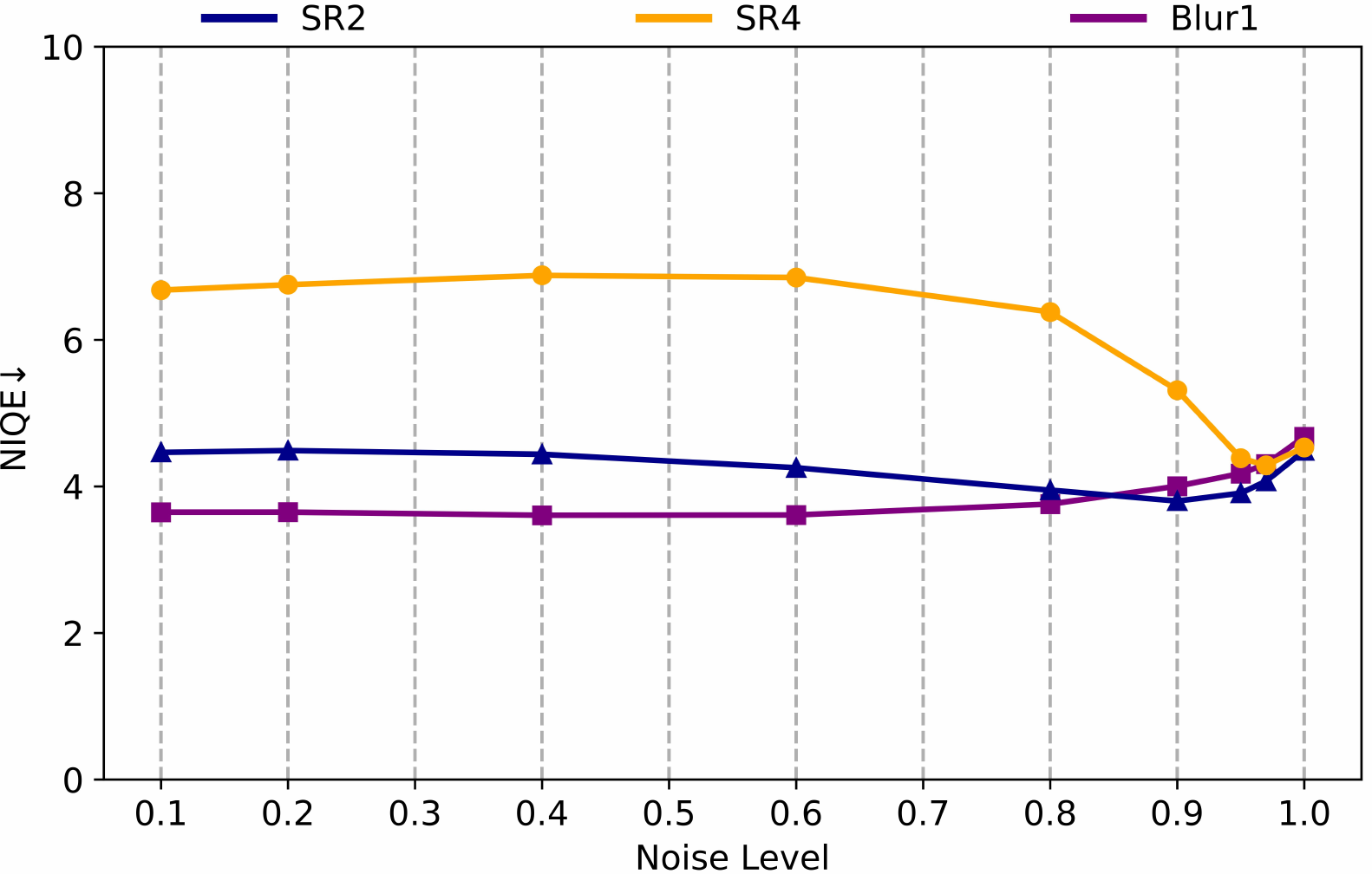}
        \caption{NIQE scores vs. noise levels}
        \label{fig:noise_perception}
    \end{subfigure}
    \caption{Impact of noise levels on distortion (SSIM) and perceptual quality (NIQE).}
    \label{fig:noise_impact}
\end{figure}

\begin{figure}[htbp]
    \centering
    \begin{subfigure}[t]{0.48\textwidth}
        \centering     
        \includegraphics[width=\textwidth]{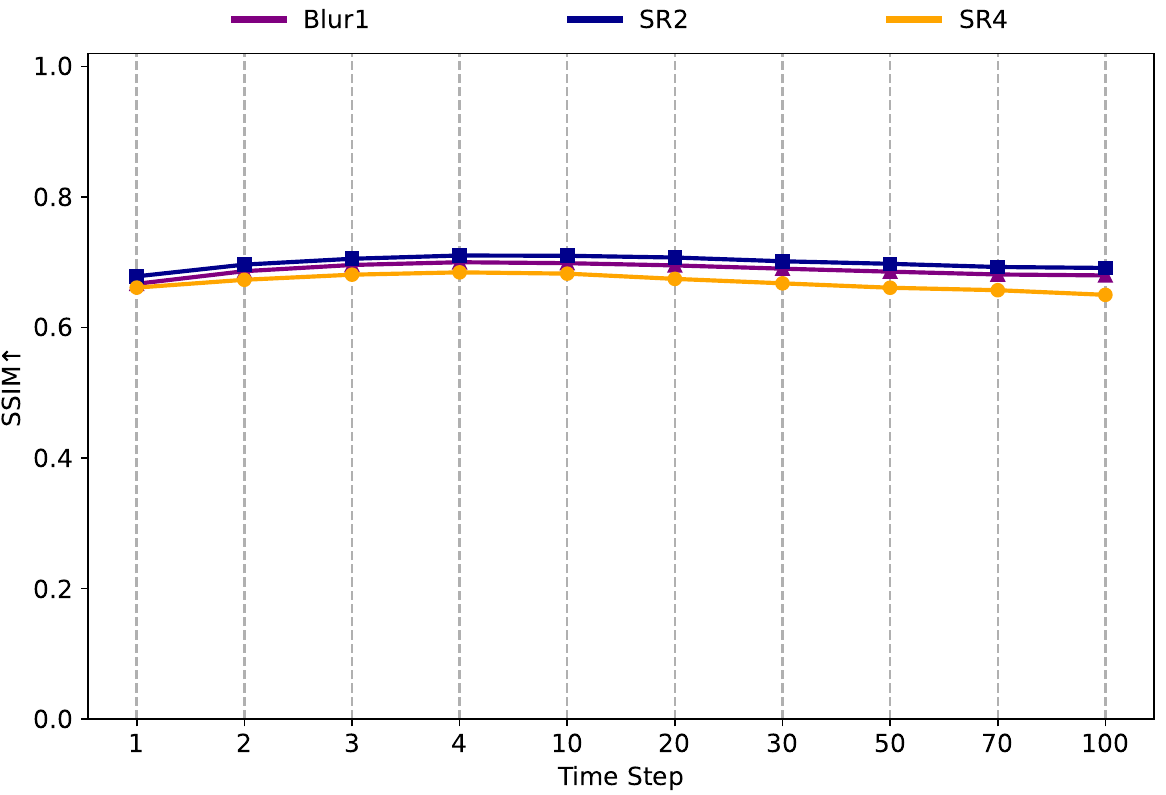}
        \caption{SSIM for different sampling steps}
        \label{fig:sampling_steps_distortion}
    \end{subfigure}
    \hfill
    \begin{subfigure}[t]{0.48\textwidth}
        \centering     
        \includegraphics[width=\textwidth]{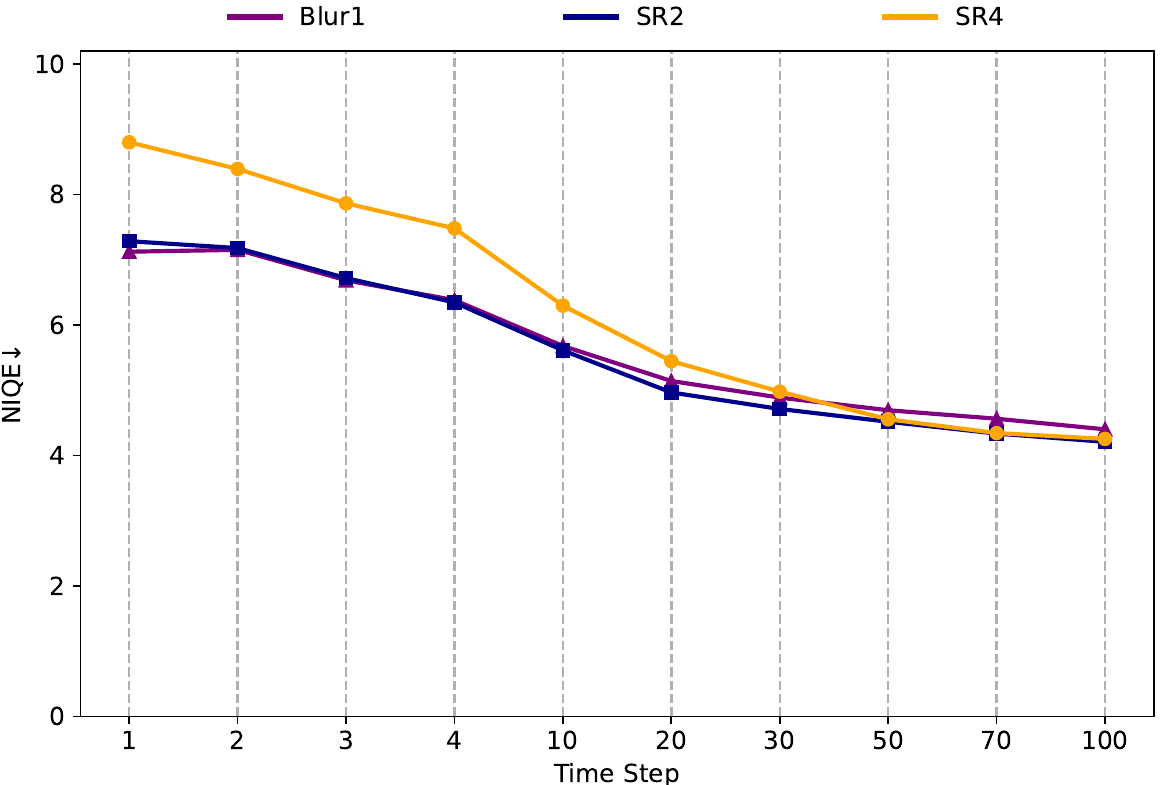}
        \caption{NIQE for different sampling steps}
        \label{fig:sampling_steps_perception}
    \end{subfigure}
    \caption{Impact of sampling steps on distortion (SSIM) and perceptual quality (NIQE).}
    \label{fig:sampling_steps_impact}
\end{figure}

\begin{figure}[htbp]
    \centering
    \begin{subfigure}[t]{0.48\textwidth}
        \centering     
        \includegraphics[width=\textwidth]{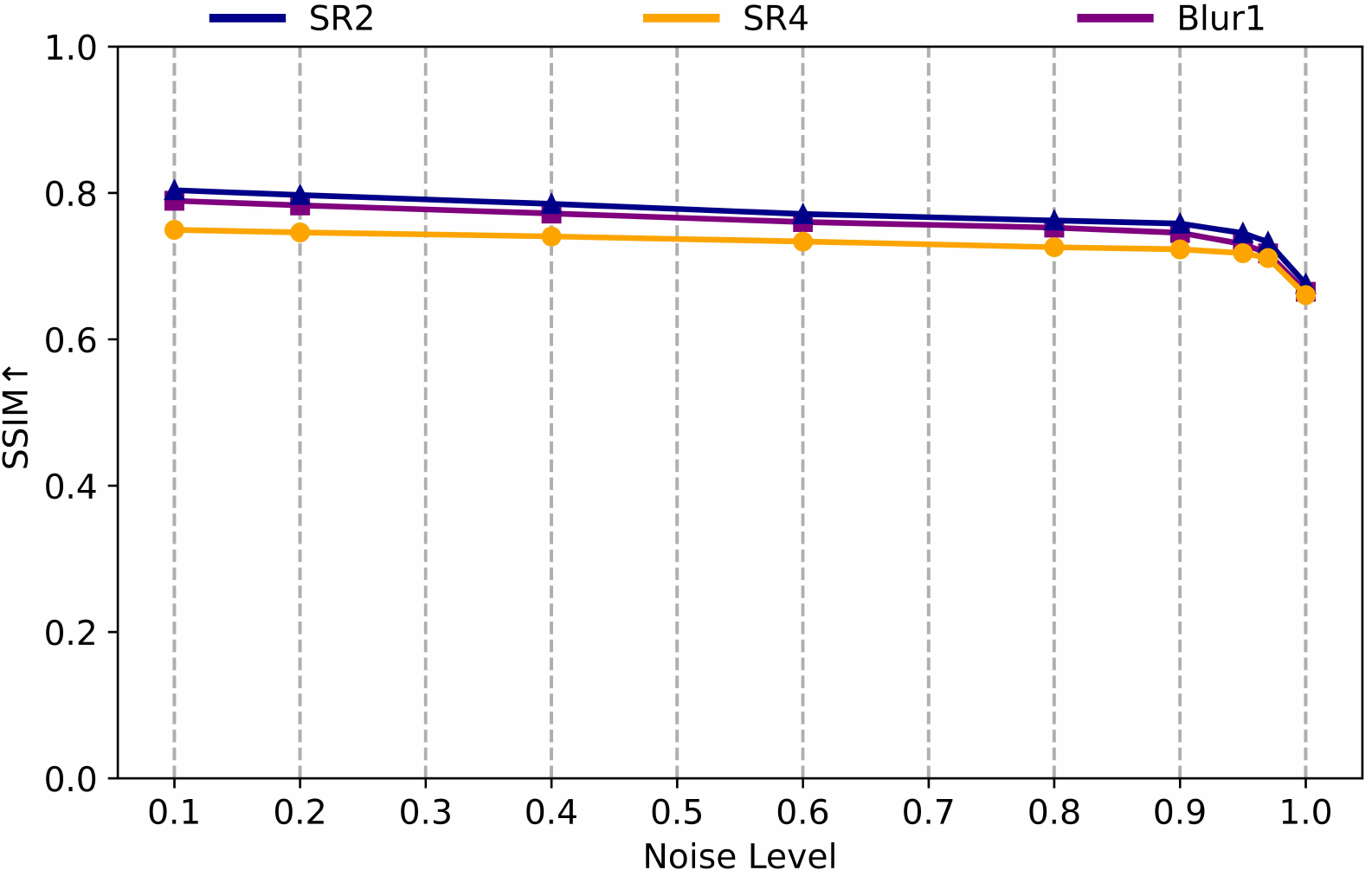}
        \caption{SSIM for different starting noise levels}
        \label{fig:one_step_distortion}
    \end{subfigure}
    \hfill
    \begin{subfigure}[t]{0.48\textwidth}
        \centering     
        \includegraphics[width=\textwidth]{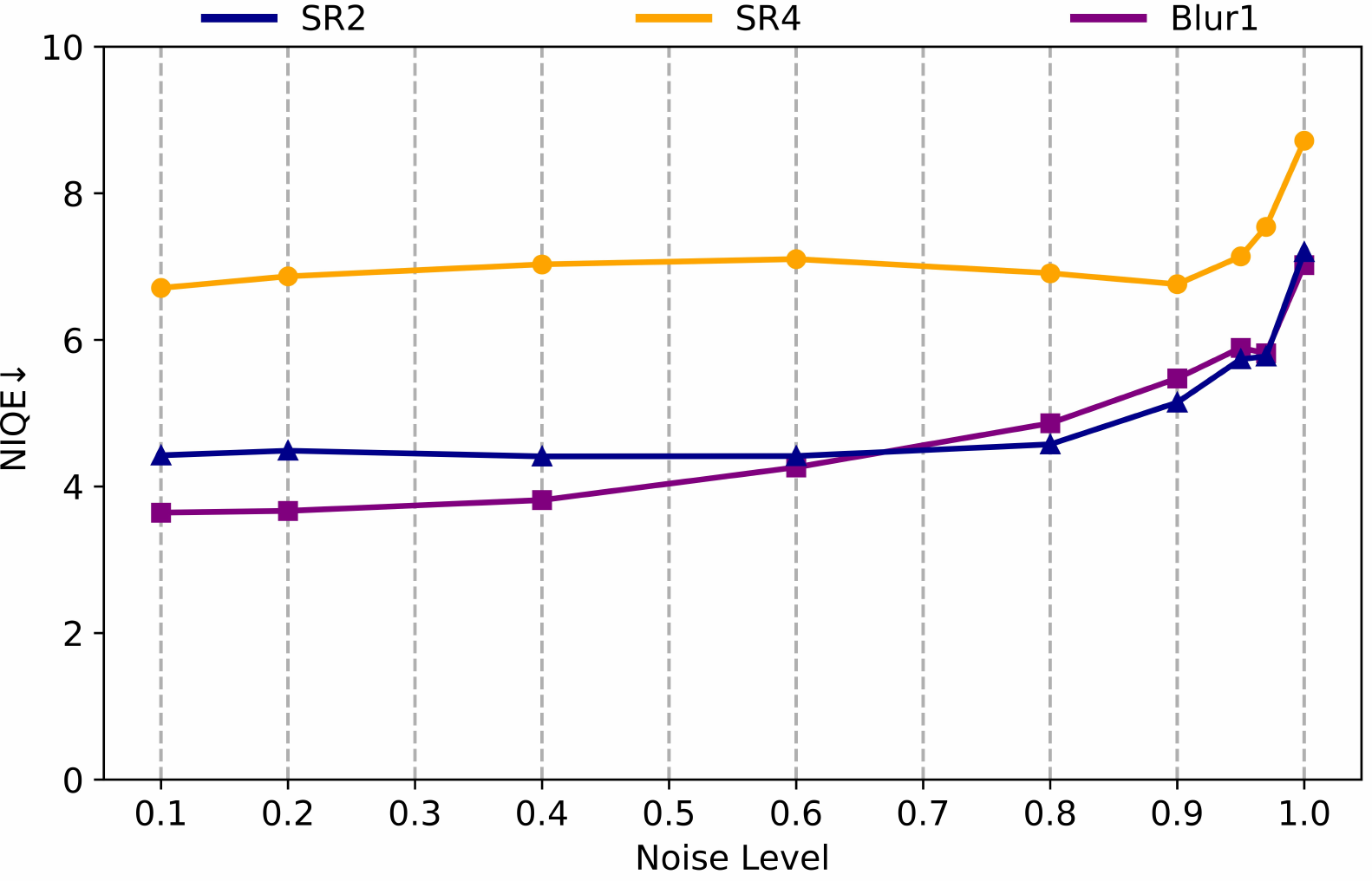}
        \caption{NIQE for different starting noise levels}
        \label{fig:one_step_perception}
    \end{subfigure}
    \caption{Impact of starting noise levels on one-step prediction.}
    \label{fig:one_step_impact}
\end{figure}

\begin{figure}[htbp]
    \centering
    \begin{subfigure}[t]{0.48\textwidth}
        \centering     
        \includegraphics[width=\textwidth]{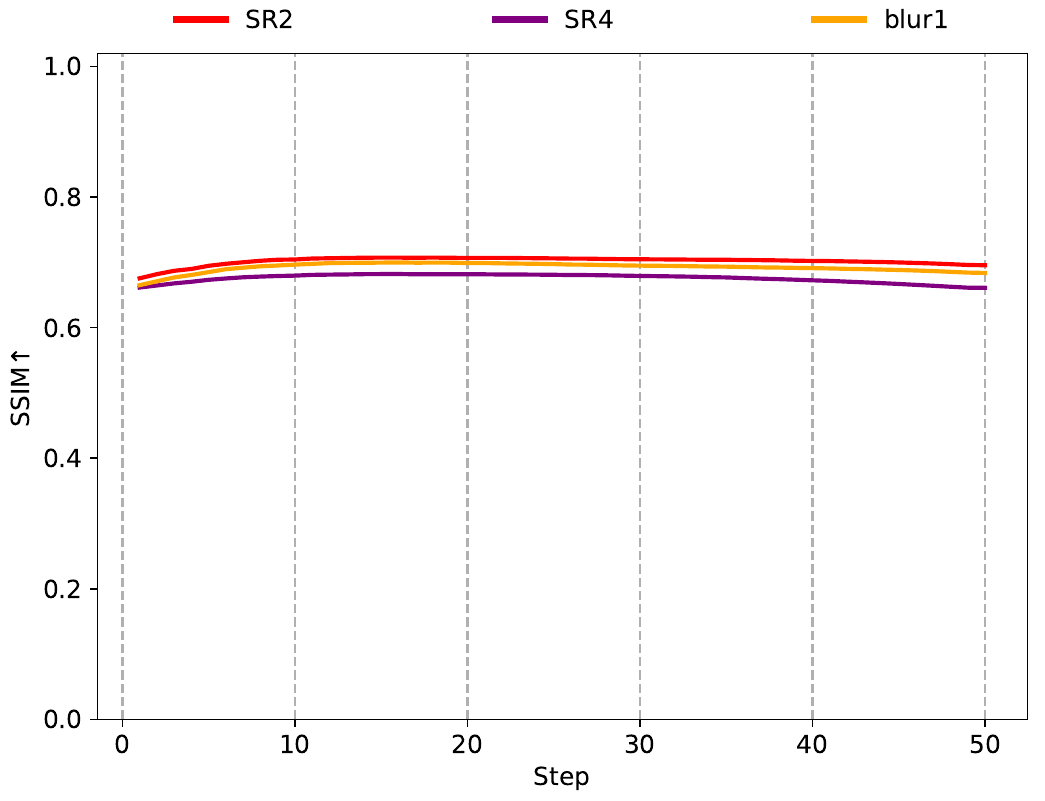}
        \caption{SSIM for different inference steps}
        \label{fig:inference_steps_distortion}
    \end{subfigure}
    \hfill
    \begin{subfigure}[t]{0.48\textwidth}
        \centering     
        \includegraphics[width=\textwidth]{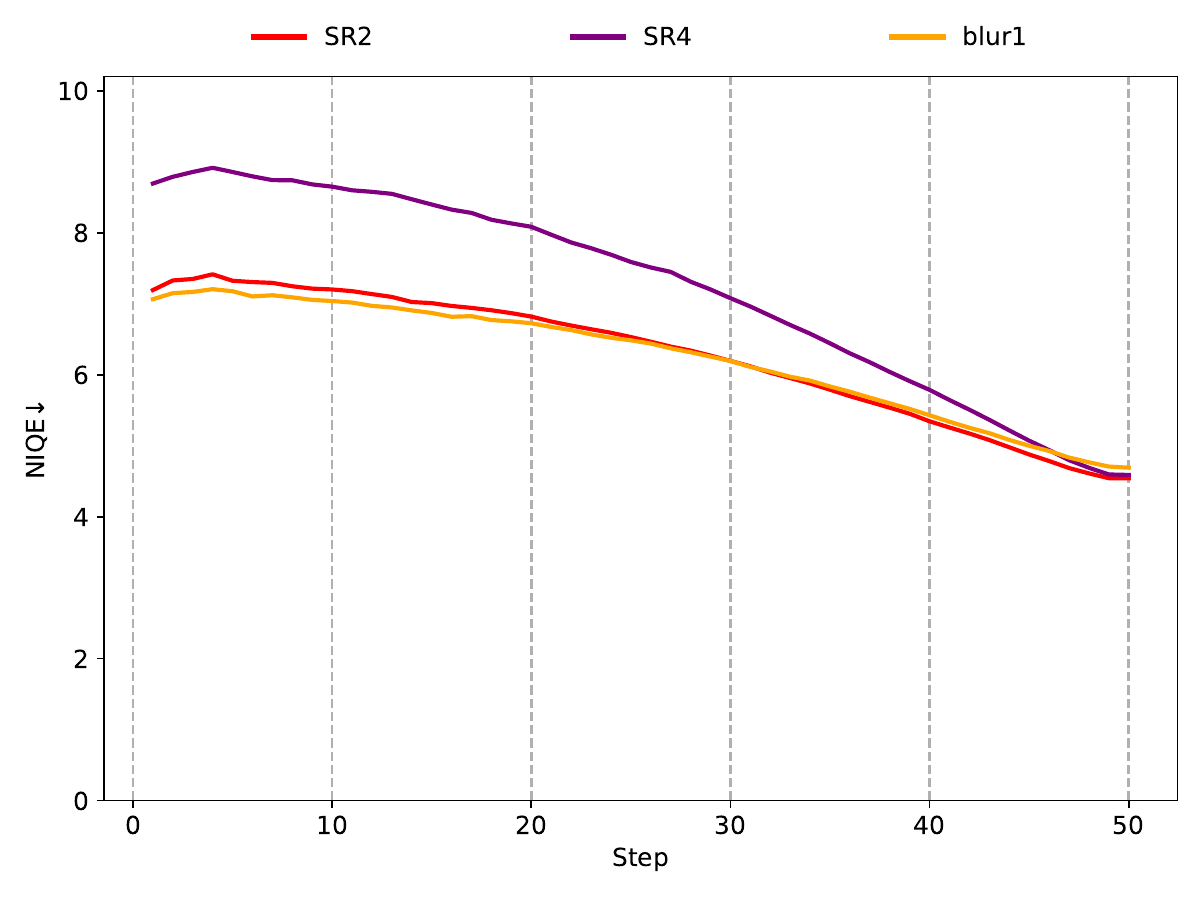}
        \caption{NIQE for different inference steps}
        \label{fig:inference_steps_perception}
    \end{subfigure}
    \caption{Impact of inference steps on distortion (SSIM) and perceptual quality (NIQE).}
    \label{fig:inference_steps_impact}
\end{figure}

Here, we provide Figure~\ref{fig:noise_impact},~\ref{fig:sampling_steps_impact},~\ref{fig:one_step_impact} and~\ref{fig:inference_steps_impact} for the main text, which are included here due to space constraints. Note that SR2 refers to 2x super-resolution, SR4 to 4x super-resolution, and Blur1 corresponds to a blur kernel with $\sigma=1$. Additionally, we include further analysis and outline future work in the supplementary material.

\section{More Analysis}

\textbf{A. Is DINOv2 Effective for Alignment Evaluation?}

To assess the semantic consistency between restored images and their degraded counterparts, we introduce \textit{alignment} as a novel evaluation aspect. Since established metrics for this purpose do not exist, we use differences in embeddings encoded by DINOv2—a visual representation learning model—to estimate semantic deviation. The concept of alignment suggests that the low-quality input and the ground truth image should demonstrate strong semantic correspondence.

As shown in Table~\ref{tab:alignment_evaluation_LQ}, calculating the DINOv2 embedding differences between the low-quality input and the ground truth image shows better results compared to other methods, even though distortion metrics fail to capture this improvement. This demonstrates that, while simple, DINOv2 embeddings effectively capture semantic information.

\begin{table}[htbp]
    \centering
    \caption{Alignment evaluation compared with ground truth images.}
    \label{tab:alignment_evaluation_LQ}
    \begin{tabular}{llcc}
        \toprule
        \textbf{Task} & \textbf{Model} & \textbf{DINOv2$\downarrow$} & \textbf{SSIM$\uparrow$} \\
        \midrule
        \multirow{3}{*}{SR2} 
        & LQ           & \textbf{0.0115} & 0.9080      \\
        & DiffBIR      & 0.5846         & 0.6987      \\
        & Stripformer  & 0.1740         & \textbf{0.9149}      \\
        \midrule
        \multirow{3}{*}{Blur1} 
        & LQ           & \textbf{0.0492} & 0.8886      \\
        & DiffBIR      & 0.5971         & 0.6907      \\
        & Stripformer  & 0.0953         & \textbf{0.9622}      \\
        \bottomrule
    \end{tabular}
\end{table}

\textbf{B. Visual Results for Alignment Evaluation}

% In this section, we present visual results for alignment evaluation. As shown in Figure~\ref{fig:task_comparison}, we observe that, across all tasks, the two CLDM-based models exhibit greater semantic deviations in details compared to traditional models. All images used in the experiments were downsampled by a factor of two using bicubic interpolation.

\begin{figure*}[htbp]
    \centering
    \includegraphics[width=0.9\textwidth]{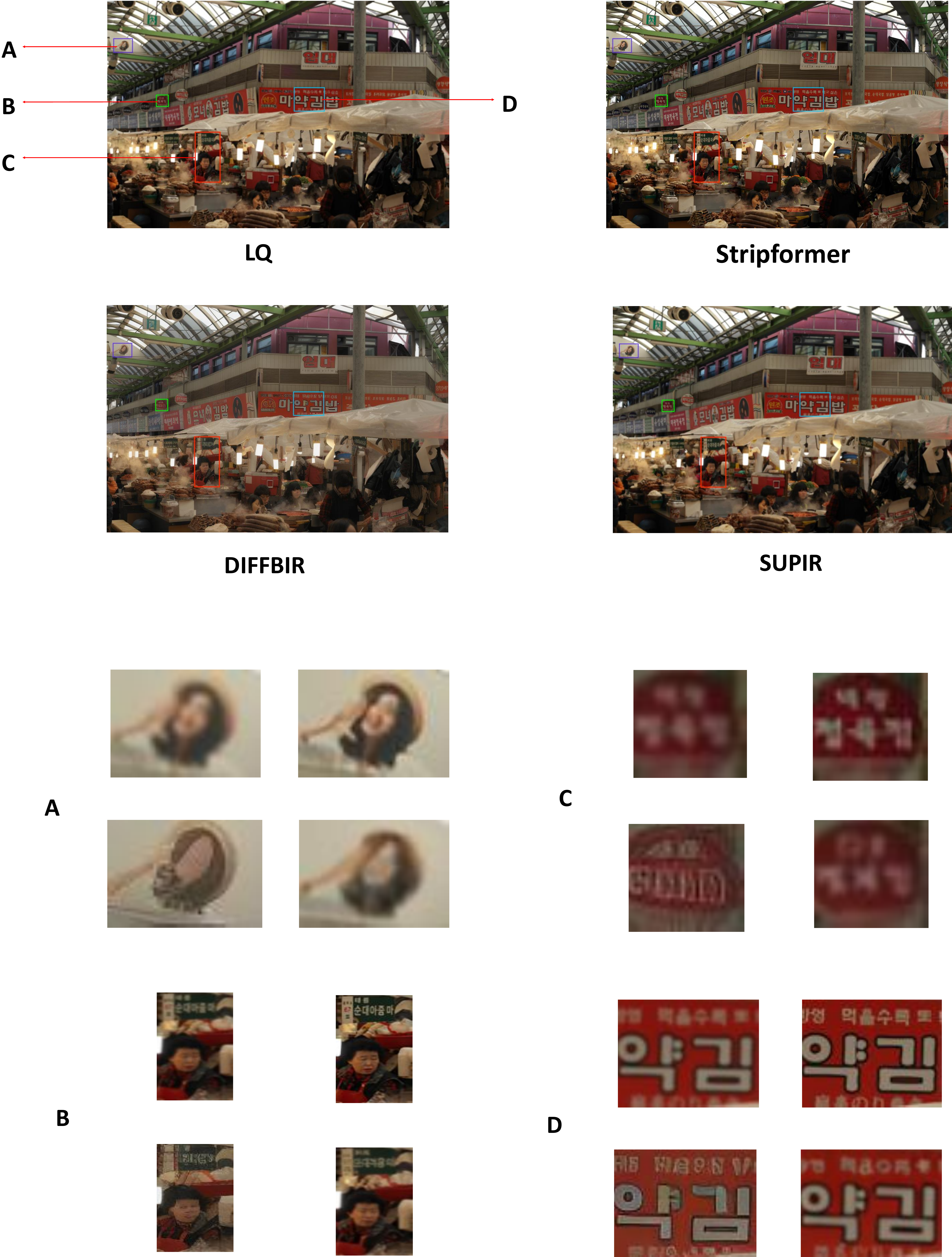}
    \vspace{0.4cm}
    \caption{Visual comparison of restored images under the \textbf{Blur1} setting. Traditional models retain better semantic consistency, while CLDM-based models exhibit more semantic deviations. Please zoom in for a better comparison.}
    \label{fig:blur1_comparison}
\end{figure*}

\begin{figure*}[htbp]
    \centering
    \includegraphics[width=0.9\textwidth]{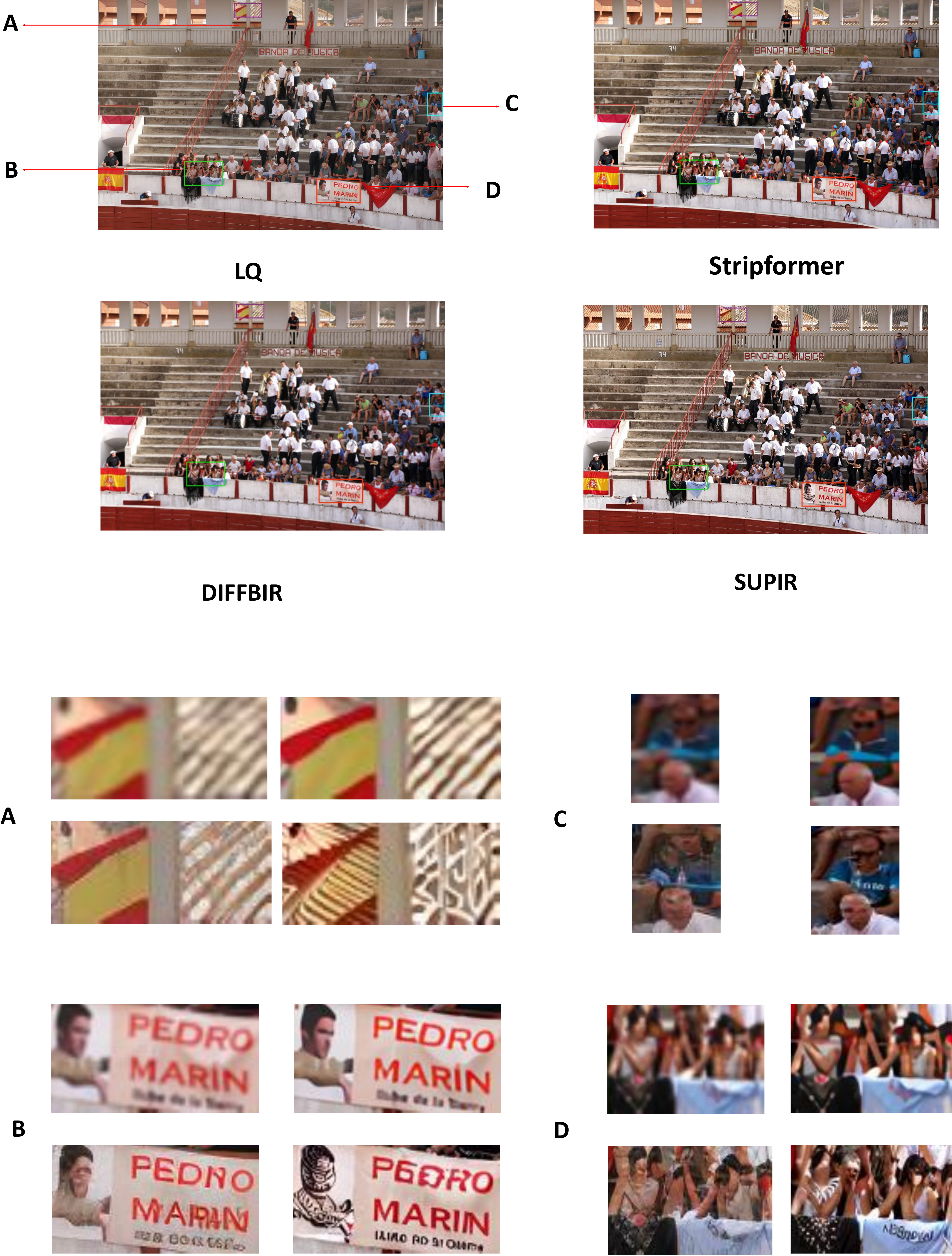}
    \vspace{0.4cm}
    \caption{Visual comparison of restored images under the \textbf{SR2} setting. The CLDM-based models struggle with fine details, resulting in greater semantic inconsistencies compared to traditional methods. Please zoom in for a better comparison.}
    \label{fig:SR2_comparison}
\end{figure*}

\begin{figure*}[htbp]
    \centering
    \includegraphics[width=0.9\textwidth]{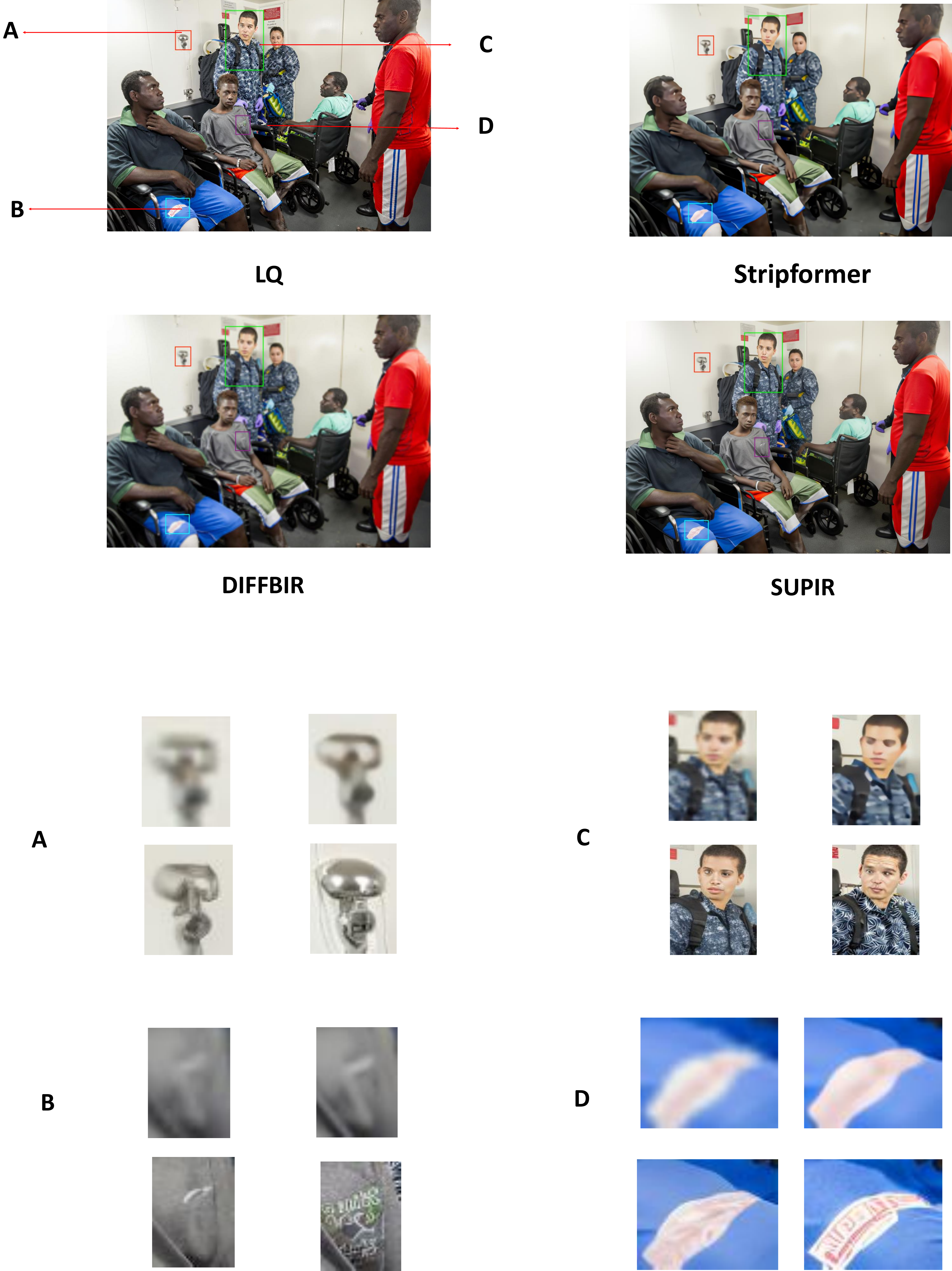}
    \vspace{0.4cm}
    \caption{Visual comparison of restored images under the \textbf{SR4} setting. Traditional models achieve better alignment, while CLDM-based models introduce noticeable semantic deviations in detailed regions. Please zoom in for a better comparison.}
    \label{fig:SR4_comparison}
\end{figure*}

In this section, we present visual results for alignment evaluation. As illustrated in Figure~\ref{fig:blur1_comparison},~\ref{fig:SR2_comparison} and ~\ref{fig:SR4_comparison}, the two CLDM-based models consistently exhibit greater semantic deviations in details compared to traditional models across all tasks. 

\section{Future Work}

We identify several areas that require further exploration:
\begin{itemize}

\item \textit{Comprehensive Evaluation of Models and Tasks.} 

The current analysis is limited to a specific set of models and tasks, which may not yield rigorous or generalizable conclusions. Future work could involve extensive testing on a broader range of models and diverse image restoration (IR) task settings, employing additional evaluation metrics. Furthermore, addressing issues of semantic deviation and model performance across different degradation levels necessitates the development of standardized benchmarking protocols.

\item \textit{In-depth Exploration of Influential Factors.} 

This study primarily investigates influential factors such as noise levels, timesteps, and latent space during inference. Future research could explore the impact of various training configurations, including alternative loss functions (e.g., in latent space, pixel space, or feature-based), dynamic noise levels, and timestep control adapted to degradation levels. Additionally, factors like restoration guidance levels, classifier-free guidance, and noise injection during sampling remain unexplored and may significantly affect performance.

\item \textit{Development of Alignment Metrics for Image Restoration.}

This work introduces alignment as a novel evaluation aspect. However, due to the absence of established metrics, it relies on preliminary methods for assessment.Future studies could focus on developing advanced metrics specifically designed to evaluate semantic consistency in IR, thereby providing novel approaches for IR evaluation.

\item \textit{Refinement of CLDM Architecture for Image Restoration.}

Our work shows a misalignment between the existing CLDM architectures and the objectives of IR tasks with various design insights on how different components affect the IR performance. Future work could work on adjusting the CLDM architecture to better fit IR tasks. We also anticipate further research efforts to address existing challenges in CLDM-based IR solutions, such as high distortion, diminished performance on samples with low levels of degradation, and semantic deviations.Given CLDM's substantial advantages in data scalability, model size, and inference efficiency, refining its architecture may unlock significant potential beyond current limitations. 

\end{itemize}

\end{document}